\documentclass{article}

\usepackage{appendix}
\usepackage{microtype}
\usepackage{graphicx}
\usepackage{subfigure}
\usepackage{booktabs} 
\usepackage{paralist}
\usepackage{url}            
\usepackage{epsfig}
\usepackage{algorithm}

\usepackage{algorithmic}
\usepackage{listings}

\usepackage{graphicx}
\usepackage{amsmath}
\usepackage{amssymb}
\usepackage{adjustbox}
\usepackage{amsthm}
\usepackage{bbm}
\usepackage{multirow}

\usepackage{threeparttable}
\usepackage{wrapfig,lipsum,booktabs}
\usepackage{xcolor, colortbl}

\usepackage[colorlinks=true, linkcolor=blue, citecolor=blue]{hyperref}

\PassOptionsToPackage{numbers, compress}{natbib}

\usepackage[final]{neurips_2021} 
\newcommand{\advCL}{\textsc{AdvCL}}
\newcommand{\name} {\advCL~}
\newcommand{\slf}{\textsc{SLF}}
\newcommand{\aff}{\textsc{AFF}}
\newcommand{\alf}{\textsc{ALF}}
\newcommand{\CFit}{\textsc{ClusterFit}}

\newcommand{\TBN}{\textsc{TriBN}}

\definecolor{Gray}{gray}{0.82}

\definecolor{GrayL}{gray}{0.92}
\definecolor{LightCyan}{rgb}{0.88,1,1}
\newcommand{\CCG}[1]{\cellcolor{Gray}}
\newcommand{\CCGL}[1]{\cellcolor{GrayL}}

\newcommand{\mycomment}[1]{}
\newcommand{\advt}{\text{AT}}

\DeclareMathAlphabet\mathbfcal{OMS}{cmsy}{b}{n}
\newcommand{\Def}[0]{\mathrel{\mathop:}=}

\DeclareMathOperator*{\argmax}{\operatorname*{argmax}}

\usepackage{pifont}
\makeatletter
\newcommand{\subalign}[1]{%
  \vcenter{%
    \Let@ \restore@math@cr \default@tag
    \baselineskip\fontdimen10 \scriptfont\tw@
    \advance\baselineskip\fontdimen12 \scriptfont\tw@
    \lineskip\thr@@\fontdimen8 \scriptfont\thr@@
    \lineskiplimit\lineskip
    \ialign{\hfil$\m@th\scriptstyle##$&$\m@th\scriptstyle{}##$\hfil\crcr
      #1\crcr
    }%
  }%
}

\makeatother

\title{When Does Contrastive Learning Preserve\\ Adversarial Robustness from\\ Pretraining to Finetuning?}
\author{Lijie Fan$^{1}$, Sijia Liu$^{2,3}$, Pin-Yu Chen$^{3}$, Gaoyuan Zhang$^{3}$, Chuang Gan$^{3}$\\
$^1$ Massachusetts Institute of Technology,
$^2$ Michigan State University,\\
$^3$ MIT-IBM Watson AI Lab, IBM Research\\
{\tt\small lijiefan@mit.edu, liusiji5@msu.edu},\\
{\tt\small \{pin-yu.chen,gaoyuan.zhang,chuangg\}@ibm.com}
}

\begin{document}

\maketitle
\begin{abstract}
Contrastive learning (CL) can learn generalizable  
feature representations  and achieve
state-of-the-art performance of downstream   tasks  
by finetuning a \textit{linear} classifier  on top of it. 
However, as adversarial robustness becomes vital in image classification,  it remains unclear whether or not
 CL is able to preserve robustness to  downstream  tasks.
The main challenge is that in the `self-supervised pretraining + supervised finetuning' paradigm, adversarial robustness is easily forgotten due to a learning task mismatch from pretraining to finetuning. We call such challenge `cross-task robustness transferability'. 
To address the above problem, 
 in this paper we revisit and advance CL principles through the lens of  robustness enhancement.  We show that (1) the design of contrastive views matters: High-frequency components of images are beneficial to improving model robustness; (2) Augmenting CL with   pseudo-supervision stimulus (e.g., resorting to feature clustering) helps preserve robustness without forgetting. 
Equipped with our new designs, we propose  {\advCL}, a novel   adversarial contrastive pretraining framework.  We show that {\advCL} is able to enhance cross-task robustness transferability  without  loss of model accuracy and finetuning efficiency.
With a thorough experimental study, 
we demonstrate that {\advCL} outperforms the state-of-the-art self-supervised robust learning   methods across multiple datasets (CIFAR-10, CIFAR-100 and STL-10) and finetuning schemes  (linear evaluation and full model finetuning). Code is available at \url{https://github.com/LijieFan/AdvCL}.
\end{abstract}

\section{Introduction}
\label{sec: intro}


Image classification has been revolutionized by  convolutional neural networks (CNNs). In spite of CNNs' generalization power, the lack of \textit{adversarial robustness}
has shown to be a main weakness that gives rise to security concerns in high-stakes applications when CNNs are applied, e.g., face recognition, medical image classification,  surveillance, and autonomous driving \cite{vakhshiteh2020threat,ma2021understanding,xu2020adversarial,lin2019defensive,cao2019adversarial}.  
The brittleness of CNNs can be easily manifested by generating tiny
input perturbations
 to completely alter the models' decision.
Such input perturbations and corresponding   perturbed  inputs are referred to as \textit{adversarial perturbations} and \textit{adversarial examples (or attacks)}, respectively \cite{Goodfellow2015explaining,carlini2017towards,papernot2016limitations,chen2017ead,xu2019structured}.

One of the most powerful  defensive schemes against adversarial attacks is adversarial training (AT) \cite{madry2017towards}, built upon a two-player game  in which an `attacker' crafts input perturbations to maximize the training objective for worst-case robustness, 
and a `defender' minimizes the maximum loss for an improved robust model against these attacks. However,
AT and  its many variants using min-max optimization \citep{kannan2018adversarial,ross2018improving,wang2019beyond,moosavi2019robustness,wang2019convergence,chen2019robust,boopathy2020proper,shafahi2019adversarial,zhang2019you,Wong2020Fast}
were restricted to supervised learning 
as  true labels of training data are required for both     supervised classifier and   attack generator (that ensures misclassification). 
The recent work \cite{zhang2019theoretically,carmon2019unlabeled,zhai2019adversarially} demonstrated that 
with a properly-designed attacker's objective, AT-type defenses can be generalized to the semi-supervised setting, and showed  that the incorporation of additional unlabeled data could further improve adversarial robustness in image classification. 
Such an extension from supervised   to semi-supervised  defenses further inspires us  to ask whether there exist \textit{unsupervised defenses} that can eliminate  the prerequisite of labeled data     but improve       model robustness.

Some  very recent literature \cite{hendrycks2019using,chen2020adversarial,jiang2020robust,kim2020adversarial,gowal2021selfsupervised} started tackling the problem of adversarial defense through the lens of self-supervised learning. Examples include augmenting a supervised task   with an unsupervised  `pretext' task for which  ground-truth label is  available  for  `free' \cite{hendrycks2019using,chen2020adversarial}, or robustifying unsupervised representation learning based only on a pretext task and then finetuning  the learned representations over downstream supervised tasks \cite{jiang2020robust,kim2020adversarial,gowal2021selfsupervised}. 
The latter scenario is of primary interest to us as a defense can then be performed  at the pretraining stage without needing any label information. Meanwhile, self-supervised
contrastive   learning (CL) 
has been outstandingly
successful in the field of representation learning: It 
can surpass a
supervised learning counterpart on  downstream image classification tasks in standard accuracy \cite{chen2020simple,grill2020bootstrap,tian2020makes,wang2020understanding,chen2020improved}.
Different from conventional self-supervised learning methods \cite{goyal2019scaling}, CL, e.g., SimCLR \cite{chen2020simple}, enforces instance discrimination by exploring
    multiple views of the same data and treating every instance under a specific view  as a class of its own \cite{purushwalkam2020demystifying}. 
    
    \begin{wrapfigure}{r}{73mm}
    \vspace{-5mm}
\begin{center}
\includegraphics[width=0.52\textwidth]{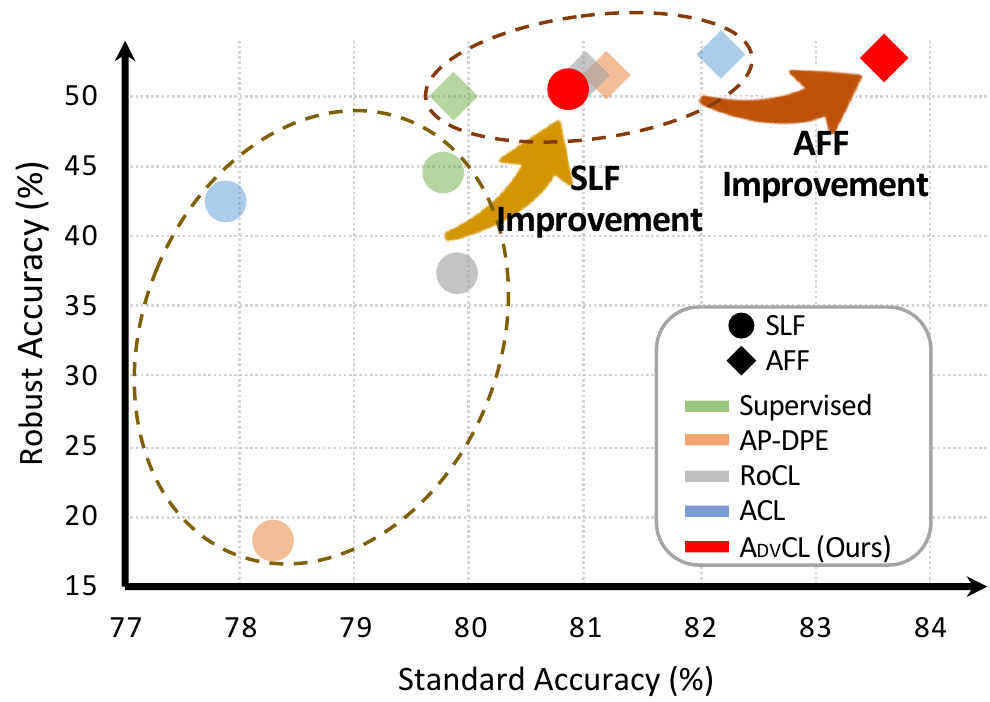}
\end{center}
\vspace{-5mm}
\caption{\footnotesize{Summary of performance for various robust pretraining methods on CIFAR-10.
The covered baseline methods include AP-DPE \cite{chen2020adversarial}, RoCL \cite{kim2020adversarial}, ACL \cite{jiang2020robust} and supervised adversarial training (AT) \cite{madry2017towards}.
Upper-right indicates better performance with respect to (w.r.t.) standard accuracy and robust accuracy (under PGD attack with $20$ steps and $8/255$ $\ell_\infty$-norm perturbation strength). Different colors represent different pretraining methods, and different shapes represent different finetuning settings. 
{Circles} (\ding{108}) indicates \textit{Standard Linear Finetuning} (SLF), and {Diamonds} (\ding{117}) indicates \textit{{Adversarial Full Finetuning}} (AFF). Our method (\advCL, red circle/diamond) has the best performance across finetuning settings. {Similar improvement could be observed under Auto-Attacks, and we provide the visualization in the appendix.}
}}
\label{fig:teaser}
\vspace*{-7mm}
\end{wrapfigure}
The most relevant work to ours is \cite{jiang2020robust,kim2020adversarial}, which integrated adversarial training with CL. However, the achieved adversarial robustness at downstream tasks largely relies on the use of advanced  finetuning techniques, either adversarial full finetuning \cite{jiang2020robust} or adversarial linear finetuning \cite{kim2020adversarial}.
Different from \cite{jiang2020robust,kim2020adversarial}, we ask:

\textit{(Q) How to accomplish robustness enhancement using  CL  without losing its finetuning efficiency, e.g., via a standard linear finetuner?}

Our work attempts to make a rigorous and comprehensive study on addressing the above question. 
We find that
self-supervised learning (including the state-of-the-art CL) suffers 
a new robustness challenge that we call `cross-task robustness transferability', which 
was largely overlooked in the previous work. 
 That is, there exists a task mismatch from pretraining to finetuning (e.g., from CL to supervised classification) so that adversarial robustness is not able to transfer across tasks even if pretraining datasets and finetuning datasets are drawn from the same distribution. 
 Different from supervised/semi-supervised learning, 
 this is a characteristic behavior of  self-supervision when being adapted to robust learning. As shown in Figure\,\ref{fig:teaser}, 
 our work advances CL in the  adversarial context and the proposed method outperforms all state-of-the-art baseline methods, leading to a substantial improvement in both robust accuracy and standard accuracy using either the lightweight standard linear finetuning or end-to-end adversarial full finetuning. 
 


\vspace{-8pt}

\paragraph{Contributions}
Our main contributions are summarized below.
    
   \ding{182} We propose  {\advCL}, a unified \underline{adv}ersarial \underline{CL} framework, and propose to use original adversarial examples and high-frequency data components to
  create  robustness-aware and generalization-aware  views of unlabeled data. 
    
\ding{183}  We propose to generate proper pseudo-supervision stimulus for  {\advCL} to improve cross-task robustness transferability. 
Different from existing self-supervised defenses aided with labeled data \cite{jiang2020robust}, we generate pseudo-labels of unlabeled data  based on their clustering information. 

    \ding{184}
    We conduct a thorough experimental study and show that
   {\advCL} achieves state-of-the-art robust accuracies   under both PGD attacks \cite{madry2017towards} and Auto-Attacks \cite{croce2020reliable} 
   using \textit{only standard linear finetuning}. For example,  in the case of Auto-Attack (the most powerful threat model) with  $8/255$  $\ell_\infty$-norm perturbation strength under ResNet-18,
   we achieve
     {$3.44\% $ and  $3.45\%$}
   robustness improvement on CIFAR-10 and CIFAR-100   over existing self-supervised methods. 
We also justify the effectiveness of {\advCL}
  in different attack setups, dataset transferring,   model explanation, and loss landscape smoothness. 

\section{Background \& Related Work}
\label{sec: background}


\textbf{Self-Supervised Learning}
Early approaches for unsupervised representation learning leverages handcrafted tasks, like prediction rotation \cite{gidaris2018unsupervised} and solving the Jigsaw puzzle \cite{noroozi2016unsupervised, carlucci2019domain}, geometry prediction~\cite{gan2018geometry} and Selfie \cite{trinh2019selfie}.
Recently contrastive learning (CL) \cite{chen2020simple,wang2020understanding,chen2020improved,oord2018representation,he2020momentum,chen2020big}
and its variants \cite{grill2020bootstrap,tian2020makes,purushwalkam2020demystifying,chen2020exploring} 
have demonstrated superior abilities in learning generalizable features in an unsupervised manner. The main idea behind CL is to self-create positive samples of the same image from aggressive viewpoints, and then
acquire  data representations  by 
maximizing agreement between positives while contrasts with negatives. 

In what follows, we elaborate on the \textbf{formulation of
SimCLR} \cite{chen2020simple}, one of the most commonly-used  
CL frameworks,    which this paper  will focus on. 
To be concrete, let $\mathcal X=\{x_1, x_2, ... , x_n\}$ denote an \textit{unlabeled source} dataset, SimCLR offers a learned
 \textit{feature encoder} $f_\theta$ to generate expressive deep representations of the data. 
 To train $f_{\boldsymbol \theta}$,
   each   input $x \in \mathcal X$ will be transformed into two  \textit{views} $(\tau_1(x), \tau_2(x))$ and labels them as a positive pair. Here transformation operations $\tau_1$ and $\tau_2$  are randomly sampled from a pre-defined transformation set  $\mathcal{T}$, which includes, e.g., random cropping and resizing, color jittering, rotation, and  cutout. The positive pair is then fed in the feature encoder $f_\theta$ with a projection head $g$ to acquire projected     features, i.e., $z_{i}=g\circ f_\theta(\tau_i(x))$ for $j \in \{ 1,2\}$. 
\textit{NT-Xent loss} (i.e., the normalized temperature-scaled cross-entropy loss) is then applied to optimizing $f_\theta$, where
 the distance of projected positive features $ (z_{1}, z_{2})$ is minimized for each input $x$.
SimCLR   follows the   `\textit{self-supervised pretraining + supervised finetuning}' paradigm. That is,
once $f_\theta$ is trained, a downstream supervised classification task can be handled by just finetuning a linear classifier $\phi$ over the fixed encoder $f_\theta$, leading to the eventual classification network $\phi\circ f_\theta$.

\textbf{Adversarial Training (AT)}
Deep neural networks are vulnerable to adversarial attacks. Various approaches have been proposed to enhance the model robustness. Given a classification model $ \theta$,
{\advt} \cite{madry2017towards} is 
one of the most powerful    robust training methods   against adversarial attacks. 
Different from standard training over normal data    $(x, y) \in \mathcal D$ (with feature $x$ and label $y$ in dataset $\mathcal D$), {\advt} adopts a min-max training recipe, where  the worst-case    training loss is minimized over the  adversarially perturbed data $(x+\delta, y)$. Here $\delta$ denotes the input perturbation variable to be maximized for the worst-case training objective. The \textit{supervised {\advt}} is then formally given by
\begin{equation}
\displaystyle
\min_{\theta}\mathbb{E}_{{(x,y)}\in{D}}\max_{\| \delta\|_\infty\leq\epsilon} \ell(x+\delta, y; \theta),
\label{eq: AT}
\end{equation}
where  $\ell$ denotes the supervised training objective, e.g., cross-entropy (CE) loss.  
There have been many variants of 
AT  \cite{shafahi2019adversarial,zhang2019you,Wong2020Fast,wong2017provable,dvijotham2018training,gan2016you,gan2016webly,zhang2019theoretically,carmon2019unlabeled,zhai2019adversarially,hendrycks2019using} established for  supervised/semi-supervised learning. 

 \vspace{-5pt}
 \paragraph{Self-supervision enabled AT}
 Several recent works \cite{chen2020adversarial,jiang2020robust,kim2020adversarial,gowal2021selfsupervised} started to study how to improve model robustness using \textit{self-supervised {\advt}}. Their  idea is to apply AT \eqref{eq: AT} to  a self-supervised pretraining task, e.g., SimCLR in \cite{jiang2020robust,kim2020adversarial},   such that  the learned feature encoder  $f_{\theta}$ renders robust data representations.
 However, different from our work, the existing ones lack a systematic study on \textit{when} and \textit{how} self-supervised robust pretraining can preserve robustness to downstream tasks without sacrificing the   efficiency of lightweight finetuning.
 For example, the prior work
 \cite{chen2020adversarial,jiang2020robust} suggested {adversarial full finetuning}, where 
 pretrained   model is used as a weight initialization in finetuning downstream tasks.
Yet, it requests the finetuner to update all of the weights of the pretrained model, and thus makes the advantage of self-supervised robust pretraining less significant. 
A more practical scenario is \textit{linear finetuning}: One freezes 
the pretrained feature encoder for the downstream task and only {partially finetunes} a linear prediction head. The work \cite{kim2020adversarial}
evaluated the performance of linear fintuning 
but
observed a relatively large performance gap between the \textit{standard} linear finetuning and \textit{adversarial} linear finetuning; see more comparisons in Figure \ref{fig:teaser}.
Therefore, the problem--\textit{how to enhance robustness transferability from pretraining to linear finetuning}--remains unexplored. 
 
 

\section{Problem Statement}
\label{sec: prob_statement}

In this section,
we present the problem of our interest, together with its setup. 



\vspace{-5pt}
\paragraph{Robust pretraining + linear finetuning.}
We aim to develop robustness enhancement solutions by fully exploiting and exploring the power of  CL 
at the pretraining phase, so that the resulting robust feature representations can seamlessly be used to generate robust predictions of downstream tasks using just a lightweight   finetuning scheme. 
With the aid of AT \eqref{eq: AT}, we formulate the `\textit{robust pretraining + linear finetuning}' problem  below:
\begin{align}
& \text{Pretraining:} ~ \displaystyle
\min_{\theta}\mathbb{E}_{{x}\in \mathcal X }\max_{\| \delta\|_\infty\leq\epsilon} \ell_{\mathrm{pre}}(x+\delta, x; \theta) \label{eq: rcl_prob} \\
&  \text{Finetuning:}~\displaystyle
\min_{\theta_{\mathrm{c}}}\mathbb{E}_{{(x,y)}\in{\mathcal D}}\, \ell_{\mathrm{CE}}( \phi_{ \theta_{\mathrm{c}}}\circ f_{\theta}(x), y),   \label{eq: slf_prob}
\end{align}
where $\ell_{\mathrm{pre}}$ denotes a properly-designed robustness- and generalization-aware CL loss (see Sec.\,\ref{sec: advCL}) given as a function of the adversarial example $(x + \delta)$, original example $x$ and   feature encoder parameters $\theta$.
In \eqref{eq: rcl_prob},
$\phi_{\theta_{\mathrm{c}}} \circ f_{\theta}$ denotes the 
classifier by equipping the linear prediction head $\phi_{\theta_{\mathrm{c}}} $ (with parameters $\theta_{\mathrm{c}}$ to be designed) on top of the fixed feature encoder $f_{\theta}$, and
$\ell_{\mathrm{CE}}$ denotes the supervised CE loss over the target dataset $\mathcal D$.
Note that besides the standard linear finetuning \eqref{eq: slf_prob}, one can also modify \eqref{eq: slf_prob} using the worst-case CE loss    for    adversarial linear/full finetuning \cite{jiang2020robust,kim2020adversarial}.  {We do not consider standard full finetuning in the paper since tuning the full network weights with standard cross-entropy loss is not possible for the model to preserve   robustness \cite{chen2020adversarial}.}

\vspace{-5pt}
\paragraph{Cross-task robustness transferability.}
Different from supervised/semi-supervised learning, self-supervision enables robust pretraining over \textit{unlabeled} source data. In the meantime, it also imposes a new challenge that we call `\textit{cross-task robustness transferability}':
At the pretraining stage, a feature encoder is learned over a ‘pretext’ task
for which ground-truth   is available for free, while finetuning is typically carried out on a new downstream   task.  
Spurred by the above, 
we ask the following questions:
\setdefaultleftmargin{1em}{1em}{}{}{}{}
\begin{compactitem}
    \item Will CL     improve adversarial robustness  using just standard linear finetuning?
    \item What  are the   principles  that    CL should follow to  preserve   robustness across tasks?
    \item 
    What are the insights can we acquire from self-supervised robust representation learning?
\end{compactitem}

\vspace{-5pt}
\section{Proposed Approach: Adversarial Contrastive Learning ({\advCL})}
\label{sec: advCL}
\vspace{-5pt}

\begin{wrapfigure}{r}{90mm}
\vspace{-7mm}
\begin{center}
\includegraphics[width=\linewidth]{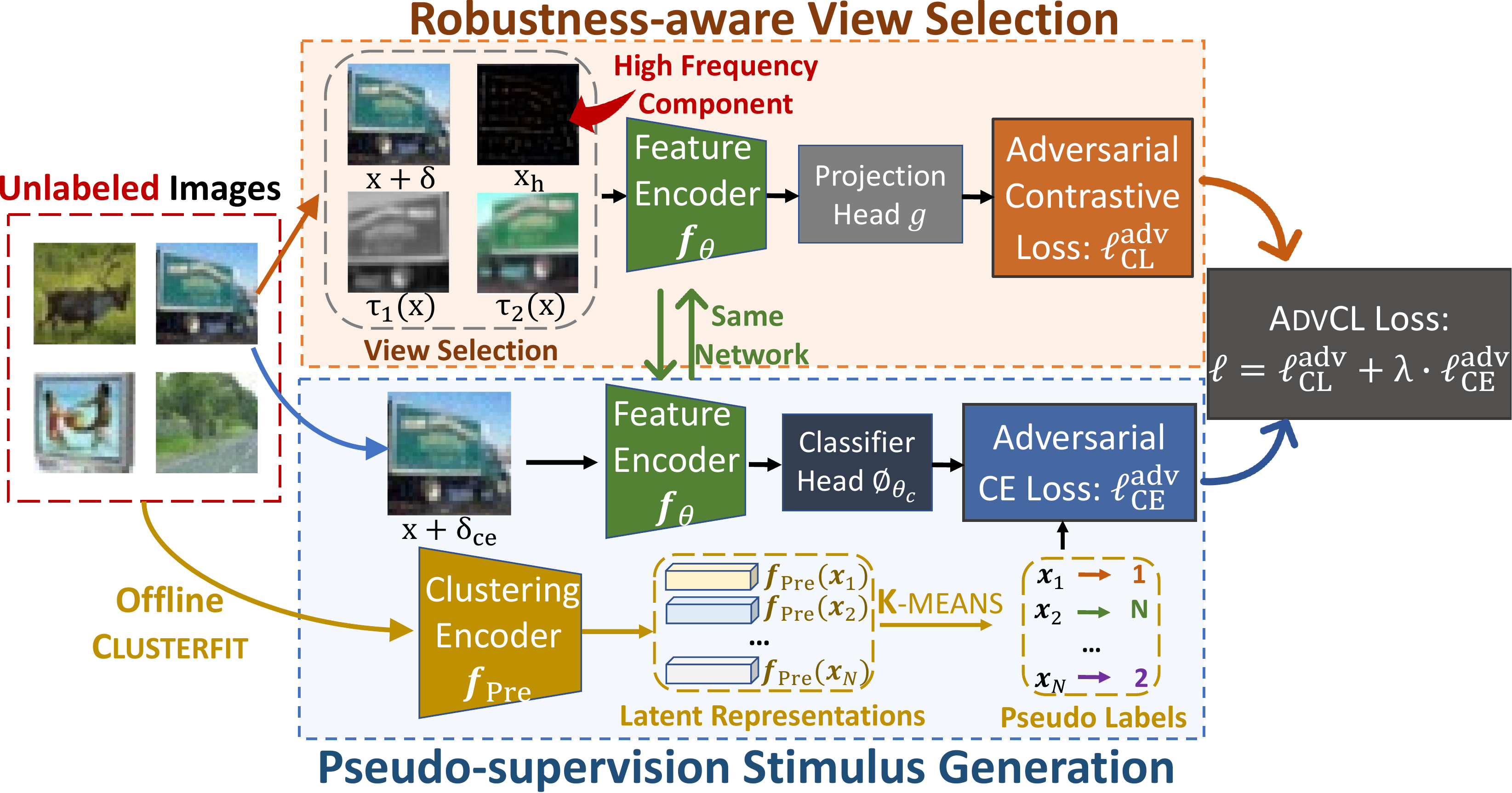}
\end{center}
\vspace{-5mm}
\caption{\footnotesize{{The overall pipeline of {\advCL}. It mainly has two ingredients: robustness-aware view selection (orange box) and pseudo-supervision stimulus generation (blue box). The view selection mechanism is advanced by high frequency components, and the supervision stimulus is created by generating pseudo labels for each image through {\textsc{ClusterFit}.
The pseudo label  (in yellow color) can be created in an offline manner and will not increase the computation overhead}.}
}}
\label{fig:model}
\vspace*{-5mm}
\end{wrapfigure}
In this section, we develop a new adversarial CL framework, {\advCL}, 
which includes two main components, robustness-aware view selection and pseudo-supervision stimulus generation.
In particular, we advance the view selection mechanism   by taking into account proper  frequency-based  data transformations that are  beneficial  to  robust  representation  learning  and  pretraining  generalization  ability.
Furthermore, we propose to  design and integrate proper supervision stimulus into {\advCL} so as to improve the cross-task robustness transferability since    robust   representations learned from self-supervision may lack the class-discriminative ability required for robust predictions on downstream tasks. 
We provide an overview of {\advCL} in Figure\,\ref{fig:model}.





\subsection{View selection mechanism}
In contrast to standard CL,
we propose two 
additional
contrastive views:    the adversarial  view and the frequency view, respectively. 

 \vspace{-5pt}
 \paragraph{Multi-view CL loss}
 Prior to defining new views, 
we first review
the  NT-Xent loss and its multi-view version used in   CL. Following notations defined in Sec.\,\ref{sec: background}, the contrastive loss with respect to (w.r.t.) a positive pair $(\tau_{1}(x), \tau_{2}(x))$ of each (unlabeled) data $x$ is given by 
%
{\begin{align}
& \ell_{\mathrm{CL}}(\tau_1(x), \tau_{2}(x))
= 
-  \sum_{i=1}^2 \sum_{j \in \mathcal P(i)} \log   \frac{\exp\big(
\text{sim}(z_{i}, z_{j})/t\big)}{\displaystyle \sum_{k\in\mathcal N(i)}  \exp\big(\text{sim}(z_{i}, z_k)/t\big)}, 
\label{eq: CL_2view}
\end{align}}%
where recall that $z_i = g\circ f(\tau_i(x))$ is the projected feature under the $i$th view, $\mathcal P(i)$ is the set of positive views except $i$  
(e.g., $\mathcal P(i)=\{ 2\}$ if $i=1$),
$\mathcal N(i)$ denotes the set of  augmented  batch data   except the  point $\tau_i(x)$, the cardinality of $\mathcal N(i)$ is $(2b-1)$ (for a data batch of size $b$ under $2$  views), 
$\text{sim}(z_{i1}, z_{i2})$ denotes the cosine similarity between representations from  two views of the same data, $\mathrm{exp}$ denotes exponential function, $\mathrm{sim}(\cdot,\cdot)$ is    the cosine similarity between two points, and $t >0$ is a temperature parameter. 
The two-view CL objective can be further extend
to the \textit{multi-view contrastive loss} \cite{khosla2020supervised}
{\begin{align}
& \ell_{\mathrm{CL}}(\tau_1(x), \tau_{2}(x), \ldots, \tau_m(x)) 
= 
-  \sum_{i=1}^m \sum_{j \in \mathcal P(i)} \log   \frac{\exp\big(
\text{sim}(z_{i}, z_{j})/t\big)}{\displaystyle \sum_{k\in\mathcal N(i)}  \exp\big(\text{sim}(z_{i}, z_k)/t\big)}, 
\label{eq: CL_mview}
\end{align}}%
 where $\mathcal P(i) = [m]/\{ i \}$ denotes the $m$ positive views except $i$, $[m]$ denotes the integer set $\{ 1,2,\ldots, m\}$, and $\mathcal N(i)$, 
 with cardinality $(bm-1)$, denotes the set of  $m$-view augmented $b$ batch samples   except the point $\tau_i(x)$.

\vspace{-5pt}
\paragraph{Contrastive view from adversarial example}
Existing methods  proposed in  \cite{jiang2020robust,kim2020adversarial,gowal2021selfsupervised} can be explained based on \eqref{eq: CL_2view}:  An adversarial perturbation $\boldsymbol \delta$ w.r.t.   each  view of   a sample $x$
is generated by maximizing the contrastive loss:
 \begin{equation}
\delta_1^*, \delta_2^*=\displaystyle \argmax_{ \| \delta_i \|_\infty\leq\epsilon } \ell_{\mathrm{CL}} (\tau_1(x)+\delta_1, \tau_2(x)+\delta_2).
\label{eq: adv_view_0}
\end{equation}
A solution to problem \eqref{eq: adv_view_0} eventually yields
 a \textit{paired} perturbation view $(\tau_1(x)+  \delta_1^*, \tau_2(x) + \delta_2^*)$.
However, the  definition of  adversarial view  \eqref{eq: adv_view_0} used in \cite{jiang2020robust,kim2020adversarial,gowal2021selfsupervised} may not be proper. 
First, standard CL commonly uses \textit{aggressive} data transformation that
treats small portions of images as positive samples of   the full image \cite{purushwalkam2020demystifying}. Despite its benefit to promoting generalization, 
crafting   perturbations over such aggressive data transformations may not be suitable for  defending  adversarial attacks applied to   \textit{full} images in the adversarial context.
Thus, a new adversarial view 
built upon   $x$ rather than $\tau_i(x)$ is desired.
Second, the contrastive loss \eqref{eq: CL_2view} is only restricted to    two views of the same data. As will be evident later, the multi-view contrastive loss is also needed when taking into account multiple robustness-promoting views.
 Spurred by above,
we   define the \textit{adversarial view} over    $x$, without modifying the existing  data augmentations $(\tau_1(x), \tau_2(x))$. This leads to the following adversarial perturbation generator by maximizing a $3$-view contrastive loss  
 \begin{equation}
\delta^* =\displaystyle \argmax_{\| \delta\|\leq\epsilon} \ell_{\mathrm{CL}} ( \tau_1(x), \tau_2(x),x+\delta),
\label{eq: atk_3view}
\end{equation}
where $x+\delta^*$ is regarded as the third view of $x$.

\vspace{-5pt}
\paragraph{Contrastive view from high-frequency component}
Next, we use the high-frequency component (HFC) of data as another additional contrastive view.
The rationale arises from the facts that 1) 
learning  over  HFC of data is a main cause of achieving superior generalization ability \cite{wang2020high} and 2) an adversary typically concentrates on HFC when manipulating an   example to fool model's decision \cite{wang2020towards}.
 Let $\mathcal{F}$ and $\mathcal{F}^{-1}$ denote Fourier transformation and its inverse. An input image $x$ can then be decomposed into
 its  HFC $x_{\mathrm h}$ and low-frequency component (LFC) $x_{\mathrm l}$:
 \begin{align}
&  x_{\mathrm h}=\mathcal{F}^{-1}(q_{\mathrm h}), \quad  x_{\mathrm l}=\mathcal{F}^{-1}(q_{\mathrm l}), \quad   [q_{\mathrm h}, q_{\mathrm l}] = \mathcal F(x).
\label{eq: LHFC}
\end{align}
In \eqref{eq: LHFC}, the distinction between $q_{\mathrm{h}}$ and $q_{\mathrm{l}}$ is made by a hard thresholding operation. Let
  $q(i,j)$ denote the $(i,j)$th element of $\mathcal F(x)$, and   $c=(c_1, c_2)$ denote the centriod of the frequency spectrum. The components $q_{\mathrm l} $ and $ q_{\mathrm h}$ in \eqref{eq: LHFC} are then generated by filtering out values according to the distance from $c$:
$
q_h(i,j)= \mathbbm{1}_{[d((i,j), (c_1,c_2))\geq r]}\cdot q(i,j) $,  and 
$ q_l(i,j)= \mathbbm{1}_{[d((i,j), (c_1,c_2))\leq r]}\cdot q(i,j)  $, 
where $d(\cdot, \cdot)$ is the Euclidian distance between two spatial coordinates, 
$r$ is a pre-defined distance threshold ($r=8$ in all our experiments), 
and $\mathbbm{1}_{[.]}\in \{0,1\}$ is an indicator function which equals to $1$ if the condition within $[\cdot]$ is met and $0$ otherwise.

\vspace{-5pt}
\paragraph{Robustness-aware contrastive learning objective}
By incorporating the adversarial perturbation $\delta$ and 
disentangling HFC $x_{\mathrm{h}}$ from the original data $x$, we   obtain a four-view contrastive loss \eqref{eq: CL_mview} 
defined over  $(\tau_1(x), \tau_2(x), x+\delta, x_{\mathrm{h}})$,
\begin{align}
\label{equ:targetCL}
  \ell_{\mathrm{CL}}^{\mathrm{adv}} (\theta; \mathcal X) \Def 
\mathbb{E}_{{x}\in{\mathcal X}}\max_{\|\delta\|_\infty\leq\epsilon} \ell_{\mathrm{CL}}( \tau_1(x), \tau_2(x),x+\delta, x_{\mathrm{h}};  \theta),
\end{align}
where recall that $\mathcal X$ denotes the unlabeled dataset, $\epsilon >0$ is a   perturbation tolerance during training, and for clarity, the four-view contrastive loss  \eqref{eq: CL_mview} is explicitly expressed as a function of model parameters $\theta$.
As will be evident latter, the eventual learning objective {\advCL} will be built upon  \eqref{equ:targetCL}.

\vspace{-5pt}
\subsection{Supervision stimulus generation: {\advCL} empowered by {\CFit}}
\vspace{-5pt}
\label{sec: CadvCL}
On top of \eqref{equ:targetCL}, we further improve the  robustness transferability of learned representations by generating a proper supervision stimulus.  
Our rationale is that robust representation could lack the class-discriminative power
required by robust classification as the former
is  acquired by optimizing an unsupervised contrastive loss while the latter is achieved by a supervised cross-entropy CE loss.
However,  there  is no  knowledge  about  supervised  data during pretraining.   
In order to improve  cross-task robustness transferability  but without calling for supervision,  
 we take advantage of {\CFit} \cite{yan2020clusterfit}, a pseudo-label generation method used in  representation learning.


To be more concrete, let $f_{\mathrm{pre}}$ denote 
a pretrained representation network that can generate latent features of unlabeled data. 
Note that $f_{\mathrm{pre}}$  can be set  available beforehand and  trained over either supervised or unsupervised dataset $\mathcal D_{\mathrm{pre}}$, e.g., ImageNet using  using CL in experiments. 
Given (normalized)
pretrained data representations $\{ f_{\mathrm{pre}}(x) \}_{x\in \mathcal X}$, {\CFit}
uses   \textit{$K$-means clustering}   to find $K$ data clusters of $\mathcal X$, and maps a \textit{cluster index} $c$   to a \textit{pseudo-label}, resulting in the pseudo-labeled dataset  $\{ (x, c) \in \hat{\mathcal X } \}$. 
By integrating {\CFit} with \eqref{equ:targetCL}, the eventual training objective of 
{\advCL} is then formed by
 \begin{align}
\label{eq: advCL_CE}
&\min_{\theta}~ \ell_{\mathrm{CL}}^{\mathrm{adv}} (\theta; \mathcal X)
+ \lambda \min_{\theta, \theta_{\mathrm{c}}}~
\underbrace{
\mathbb{E}_{{(x,c)}\in{\hat{\mathcal X}}
} \max_{\|\delta_{ce}\|_\infty\leq\epsilon
} \ell_{\mathrm{CE}}(\phi_{ \theta_{\mathrm{c}}} \circ f_{\theta} (x+\delta_{ce}),c)}_\text{Pseudo-classification enabled AT regularization},
\hspace*{-2mm}
\end{align}
where $\hat{\mathcal X}$ denotes the pseudo-labeled dataset of $\mathcal X$, $\phi_{\theta_{\mathrm c}}$ denotes a prediction head over $f_{\theta}$,   and
$\lambda > 0$ is a regularization parameter that strikes a balance between  adversarial contrastive training and pseudo-label stimulated AT. 
{When the number of clusters $K$ is not known \textit{a priori},  we   extend \eqref{eq: advCL_CE} to an \textit{ensemble version}  over $n$ choices of cluster numbers $\{ K_1, \ldots, K_n\}$.  Here each cluster number $K_i$ is paired with a unique linear classifier $\phi_i$ to obtain the supervised prediction $\phi_i\circ f$ (using cluster labels). The ensemble CE loss, given by the average of   $n$ individual losses, is then used in \eqref{eq: advCL_CE}. Our experiments show that the ensemble version usually leads to better generalization ability. 
}

\vspace{-5pt}
\section{Experiments}
\label{sec: exp}
\vspace{-5pt}

In this section, 
 we     
 demonstrate the effectiveness of our proposed 
 {\advCL}
from the following aspects: 
(1) Quantitative results, including cross-task robustness transferability, cross-dataset robustness transferability, and   robustness against PGD attacks \cite{madry2017towards} and Auto-Attacks \cite{croce2020reliable};
(2) Qualitative results, including representation t-SNE \cite{van2008visualizing}, feature inversion map visualization, and geometry of loss landscape;
(3) Ablation studies of   {\advCL}, including   finetuning schemes, view selection choices, and supervision stimulus variations. 

\vspace{-5pt}
\paragraph{Experiment setup}
 We consider three robustness evaluation metrics: (1) \underline{A}uto-attack \underline{a}ccuracy (\textbf{AA}), namely, classification accuracy over adversarially perturbed images via Auto-Attacks; (2) \underline{R}obust \underline{a}ccuracy (\textbf{RA}), namely, classification accuracy over adversarially perturbed images via PGD attacks; and (3)  \underline{S}tandard \underline{a}ccuracy (\textbf{SA}), namely,   standard classification accuracy over benign images without perturbations.
We use ResNet-18 for the encoder architecture of $f_{\boldsymbol \theta}$ in CL.
Unless specified otherwise, we use $5$-step $\ell_{\infty}$ projected gradient descent (PGD) with $\epsilon = 8/255$ to generate perturbations during pretraining, and use Auto-Attack and $20$-step $\ell_{\infty}$ PGD  with $\epsilon = 8/255$    to generate perturbations in computing AA and RA at test time. 
We will compare {\advCL} with the CL-based adversarial pretraining \textbf{baselines}
, {ACL} \cite{jiang2020robust}, RoCL \cite{kim2020adversarial}, (non-CL) self-supervised adversarial learning baseline {AP-DPE} \cite{chen2020adversarial} and the supervised AT baseline \cite{madry2017towards}.

\begin{table}[htp]
\begin{center}
\caption{\small{Cross-task performance of \name (in dark gray color), compared with supervised (in white color) and self-supervised (in light gray color) baselines, in terms of AA, RA and SA  on CIFAR-10 with ResNet-18. The pretrained models are evaluated under the standard linear finetuning (\slf) setting and the adversarial full finetuning (\aff) setting.
The top performance is highlighted in \textbf{bold}.
}}
\vspace*{1.5mm}
\label{table: overall}
\begin{threeparttable}
\resizebox{0.9\textwidth}{!}{
\begin{tabular}{c|c|c|c|c|c|c|c} 
\toprule[1.2pt]\toprule
\multirow{2}{*}{\begin{tabular}[c]{@{}c@{}}Pretraining\\Method\end{tabular}} & \multirow{2}{*}{\begin{tabular}[c]{@{}c@{}}Finetuning\\Method\end{tabular}}               & \multicolumn{3}{c|}{CIFAR-10} & \multicolumn{3}{c}{CIFAR-100}  \\ 
\cmidrule{3-8}
                                                                             &                                                                                           & AA(\%)    & RA(\%)  & SA(\%)            & AA(\%)     & RA(\%)  & SA(\%)             \\ 
\midrule
Supervised                                                                       & \multirow{5}{*}{\begin{tabular}[c]{@{}c@{}}Standard\\linear\\finetuning\\\bf(\slf)\end{tabular}} &  42.22 & 44.4  & 79.77        & 19.53  &23.41   & \bf 50.53         \\
\CCGL{}AP-DPE\cite{chen2020adversarial}                                                                          &                                                                                           & \CCGL{}16.07 & \CCGL{}18.22  & \CCGL{}78.30        & \CCGL{}4.17  & \CCGL{}6.23   & \CCGL{}47.91         \\
\CCGL{}RoCL\cite{kim2020adversarial}                                                                          &                                                                                           & \CCGL{}28.38 & \CCGL{}39.54  & \CCGL{}79.90        & \CCGL{}8.66  & \CCGL{}18.79  & \CCGL{}49.53         \\
\CCGL{}ACL\cite{jiang2020robust}                                                                          &                                                                                           & \CCGL{}39.13 & \CCGL{}42.87  & \CCGL{}77.88        & \CCGL{}16.33 & \CCGL{}20.97  & \CCGL{}47.51         \\
\CCG{}\bf \name(ours)                                                                         &                                                                                           & \CCG{}\bf 42.57 & \CCG{}\bf 50.45  & \CCG{}\bf 80.85        & \CCG{}\bf 19.78 & \CCG{}\bf 27.67  &\CCG{}48.34         \\
\midrule
Supervised                                                                       & \multirow{5}{*}{\begin{tabular}[c]{@{}c@{}}Adversarial\\full\\finetuning\\\bf(\aff)\end{tabular}} &  46.19 & 49.89  & 79.86        & 21.61  &25.86   & 52.22         \\
\CCGL{}AP-DPE\cite{chen2020adversarial}                                                                          &                                                                                           & \CCGL{}48.13 & \CCGL{}51.52 & \CCGL{}81.19    & \CCGL{}22.53  & \CCGL{}26.89  & \CCGL{}55.27         \\
\CCGL{}RoCL\cite{kim2020adversarial}                                                                          &                                                                                           & \CCGL{}47.88 & \CCGL{}51.35  & \CCGL{}81.01        & \CCGL{}22.38  & \CCGL{}27.49  & \CCGL{}55.10         \\
\CCGL{}ACL\cite{jiang2020robust}                                                                          &                                                                                           & \CCGL{}49.27 & \CCGL{}\bf 52.82  & \CCGL{}82.19        & \CCGL{}23.63 & \CCGL{}\bf 29.38  & \CCGL{}56.61         \\
\CCG{}\bf \name(ours)                                                                         &                                                                                           & \CCG{}\bf49.77 & \CCG{}52.77  & \CCG{}\bf 83.62        & \CCG{}\bf 24.72 & \CCG{}28.73  & \CCG{}\bf 56.77         \\
\bottomrule\bottomrule[1.2pt]
\end{tabular}}

\end{threeparttable}
\end{center}
\vspace{-4mm}
\vspace*{-0mm}
\end{table}

\subsection{Quantitative Results}
\vspace{-5pt}
\paragraph{Overall performance from pretraining to finetuning (across tasks)}
In Table\,\ref{table: overall},
we  evaluate the   robustness of a  classifier (ResNet-18)
finetuned over
  robust representations 
learned by different supervised/self-supervised pretraining approaches over CIFAR-10 and CIFAR-100.
We focus on two  representative finetuning schemes: the simplest standard linear finetuning (\slf) and the end-to-end adversarial full finetuning (\aff).
As we can see, the proposed {\advCL} method yields a  substantial improvement
over almost all     baseline methods. Moreover, {\advCL} improves robustness and standard accuracy simultaneously. 


\begin{wraptable}{r}{95mm}
 \vspace{-8mm}
\begin{center}
\caption{\small{Cross-dataset performance of \name (dark gray color), compared with supervised (white color) and self-supervised (light gray) baselines, in AA, RA, SA,  on STL-10 with ResNet-18.}}
\vspace{2mm}
\label{table: transfer}
\begin{threeparttable}
\resizebox{0.68\textwidth}{!}{
\begin{tabular}{l|c|c|c|c|c|c|c}
\toprule[1.2pt]\toprule
\multirow{2}{*}{Method} & \multirow{2}{*}{
\begin{tabular}[l]{@{}l@{}}Fine-\\tuning\end{tabular}
}    & \multicolumn{3}{c|}{CIFAR-10 $\xrightarrow[]{}$ STL-10} & \multicolumn{3}{c}{CIFAR-100 $\xrightarrow[]{}$ STL-10}  \\ 
\cmidrule{3-8}
                        &                                        & AA(\%)    & RA(\%)    & SA(\%)            & AA(\%)    & RA(\%)    & SA(\%)             \\ 
\midrule
Supervised          & \multirow{4}{*}{\begin{tabular}[c]{@{}c@{}}\bf\slf\end{tabular}}                 & 22.26 & 30.45 & 54.70      & 19.54 & 23.63 & \bf51.11          \\
\CCGL{}RoCL\cite{kim2020adversarial}                                            &                         & \CCGL{}18.65 & \CCGL{}28.18 & \CCGL{}54.56         & \CCGL{}12.39 & \CCGL{}21.93 & \CCGL{}47.86          \\
\CCGL{}ACL\cite{jiang2020robust}                                            &                         & \CCGL{}25.29 & \CCGL{}31.80 & \CCGL{}55.81         & \CCGL{}\bf21.75 & \CCGL{}26.32 & \CCGL{}45.91          \\
\CCG{}\bf \name(ours)                                                         &                         & \CCG{}\bf25.74 & \CCG{}\bf35.80 & \CCG{}\bf63.73         & \CCG{}20.86 & \CCG{}\bf30.35 & \CCG{}50.71          \\ 
\midrule
Supervised          & \multirow{4}{*}{\begin{tabular}[c]{@{}c@{}}\bf\aff\end{tabular}}                                          & 33.10 & 36.7  & 62.78         & 29.18 & 32.43 & 55.85          \\
\CCGL{}RoCL\cite{kim2020adversarial}                                                &                         & \CCGL{}29.40 & \CCGL{}34.65 & \CCGL{}61.75         & \CCGL{}27.55 & \CCGL{}31.38 & \CCGL{}57.83          \\
\CCGL{}ACL\cite{jiang2020robust}                                           &                         & \CCGL{}32.50 & \CCGL{}35.93 & \CCGL{}62.65         & \CCGL{}28.68 & \CCGL{}32.41 & \CCGL{}57.16          \\
\CCG{}\bf \name(ours)                                                          &                         & \CCG{}\bf34.70 & \CCG{}\bf37.78 & \CCG{}\bf63.52         & \CCG{}\bf30.51 & \CCG{}\bf33.70 & \CCG{}\bf61.56         \\
\bottomrule\bottomrule[1.2pt]
\end{tabular}}

\end{threeparttable}
\end{center}
\vspace{-5mm}
\end{wraptable}
\vspace{-5pt}
\paragraph{Robustness transferability across datasets}
In Table\,\ref{table: transfer}, we next evaluate the robustness transferability across different datasets, where $A \to B$ denotes the transferability from pretraining on dataset $A$ to finetuning on another dataset $B$ ($\neq A$)
of representations learned by {\advCL}.
Here the pretraining setup is consistent with Table\,\ref{table: overall}.
We observe that {\advCL} yields  better robustness as well as standard accuracy than almost all   baseline approaches   under both {\slf} and {\aff} finetuning settings.  In the case of CIFAR-100 $\xrightarrow[]{}$ STL-10, although {\advCL}  yields $0.89\%$ AA drop compared to ACL \cite{jiang2020robust}, it yields a  much better SA with $4.8\%$   improvement.

\vspace{-5pt}
\paragraph{Robustness evaluation vs. attack strength}
\begin{wrapfigure}{r}{87mm}
\begin{center}
\vspace*{-8mm}
\includegraphics[width=\linewidth]{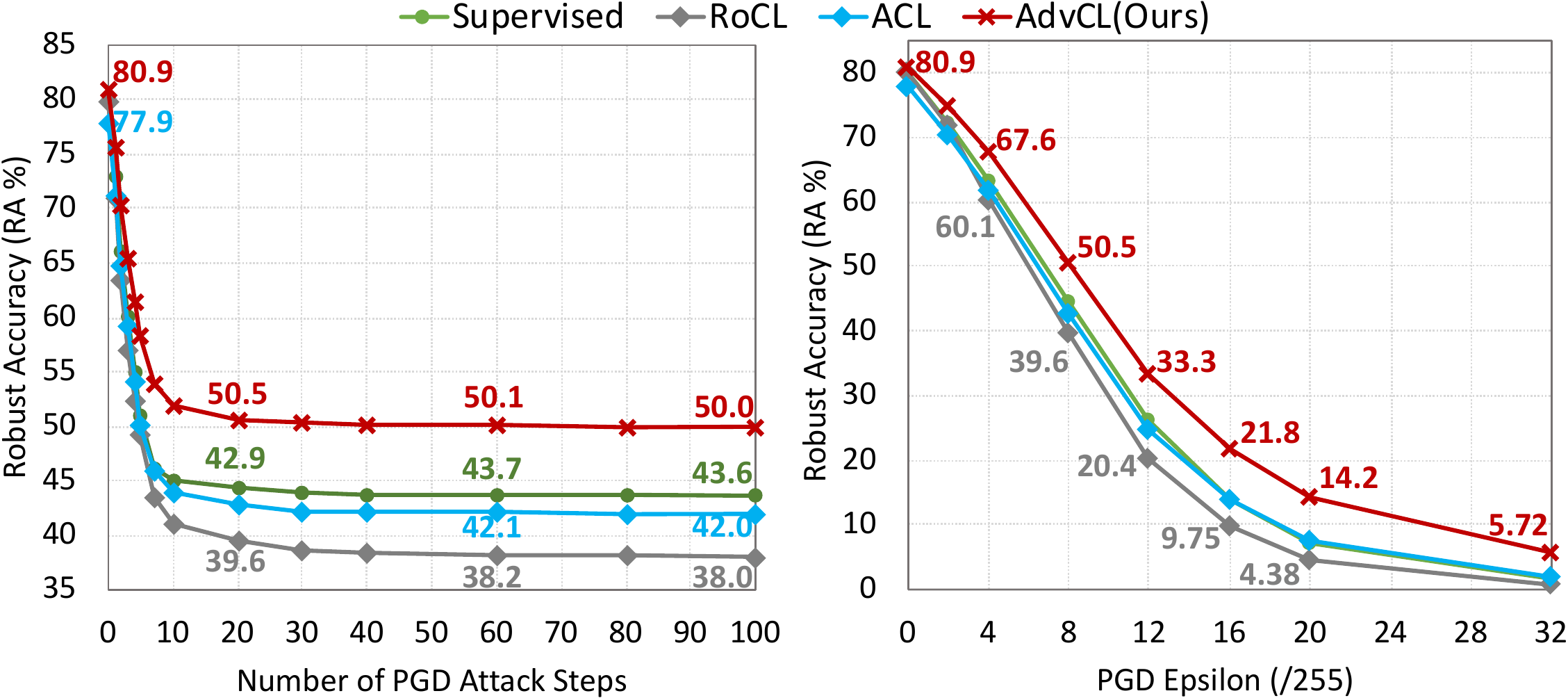}
\end{center}
\vspace*{-3mm}
\caption{RA of {\advCL} and baseline approaches under various PGD attacks. \slf~is applied to the pretrained model.
}
\label{fig:PGD_atk}
\vspace{-3pt}
\end{wrapfigure}

It was shown in  \cite{athalye2018obfuscated}  that 
an  adversarial defense that causes \textit{obfuscated} gradients results in a \textit{false sense of model robustness}. The issue of obfuscated gradients typically comes with two `side effects': (a) The success rate of PGD attack   {ceases} to be improved as the $\ell_\infty$-norm perturbation radius $\epsilon$ increases; (b) A larger number of PGD steps {fails} to generate stronger adversarial examples. 
Spurred by the above, 
Figure\,\ref{fig:PGD_atk} shows the finetuning performance of {\advCL} (using  {\slf}) as a function of 
the perturbation size $\epsilon$ and the PGD  step number. As we can see, {\advCL} is consistently more robust than the baselines at all  different PGD settings for a significant margin.
\subsection{Qualitative Results}

\begin{figure}[htb]
\begin{center}
\vspace*{-2mm}
\begin{tabular}{ccc}
\includegraphics[width=0.3\textwidth]{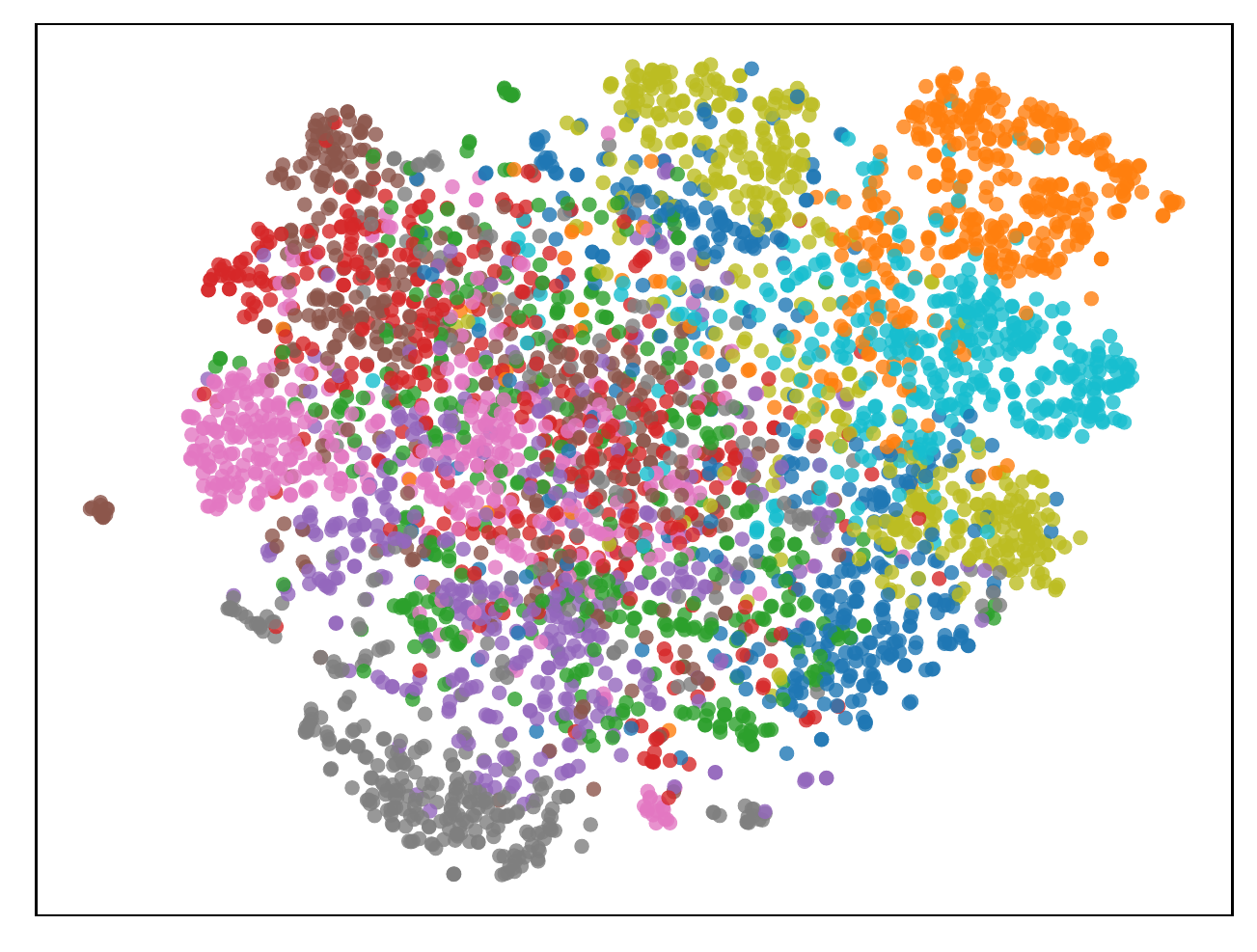} & \hspace*{-0.12in}
\includegraphics[width=0.3\textwidth]{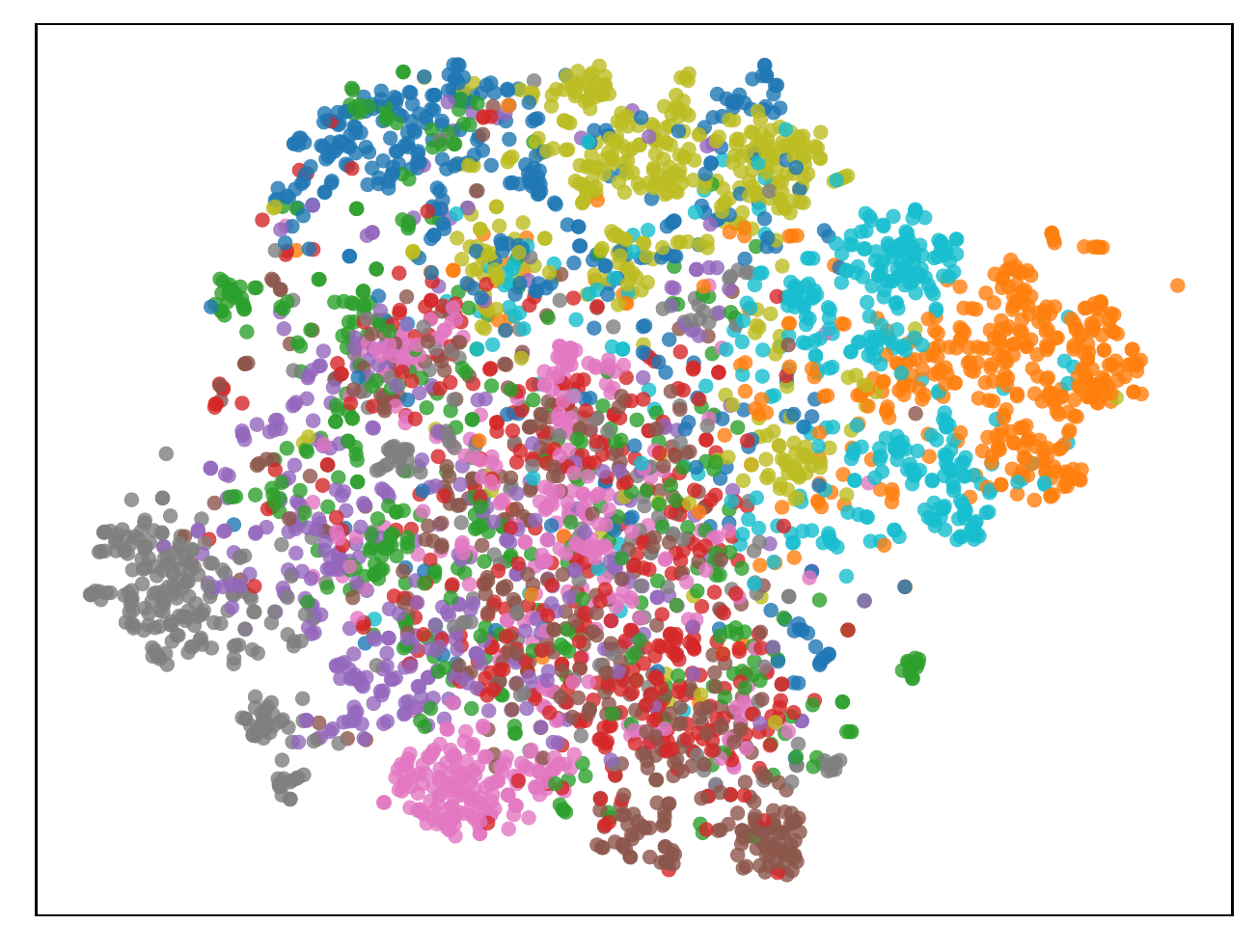} & \hspace*{-0.12in}
\includegraphics[width=0.3\textwidth]{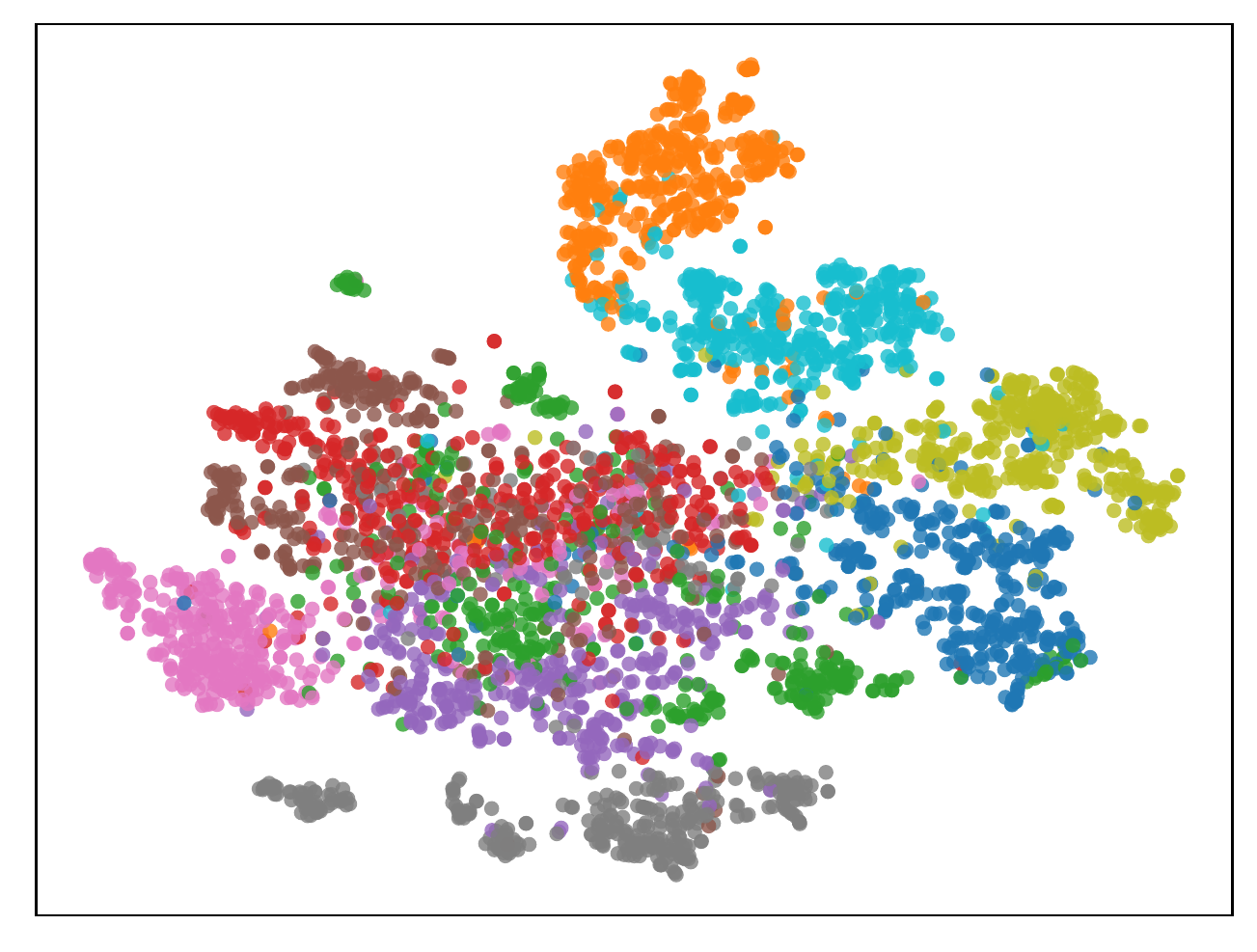}
\\
{\footnotesize (a) RoCL \cite{kim2020adversarial}} & \hspace*{-0.12in} {\footnotesize (b) ACL \cite{jiang2020robust}} & \hspace*{-0.12in} {\footnotesize (c) {\advCL} \bf(ours)} 
\end{tabular}
\end{center}
\vspace*{-3mm}
\caption{t-SNE visualization of representations learned with different self-supervised pretraining approaches. Our {\advCL} gives a much clearer separation among classes than baseline approaches.
}
\label{fig:tsne}
\vspace{-5pt}
\end{figure}

\vspace{-5pt}
\paragraph{Class discrimination of learned representations}
To further demonstrate the efficacy of \advCL, Figure\,\ref{fig:tsne} visualizes the representations learned by self-supervision using t-SNE \cite{van2008visualizing} on CIFAR-10. We color each point using its ground-truth label. The results show representations learned by {\advCL}  have a much clearer class boundary than those learned with baselines. This indicates that    {\advCL} makes an adversary difficult to successfully perturb an image, leading to a more robust prediction.

\begin{wrapfigure}{r}{67mm}
 \vspace*{-7mm}
  \centering
  \begin{adjustbox}{max width=0.4\textwidth }
  \begin{tabular}{@{\hskip 0.00in}c @{\hskip 0.01in} | @{\hskip 0.01in}c   @{\hskip 0.00in}   @{\hskip 0.00in} c @{\hskip 0.00in}   @{\hskip 0.00in} c @{\hskip 0.00in}  @{\hskip 0.0in} c @{\hskip 0.00in}    @{\hskip 0.0in} c
  }
\colorbox{GrayL}{ \textbf{\Large{Seed Images}}} 
&
\textcolor{blue}{ \textbf{\Large{AT}}}
&

\textcolor{blue}{\Large{ \textbf{RoCL}} }
& 

\textcolor{blue}{\Large{  \textbf{ACL}}}
& 
\textcolor{blue}{ \Large{  \textbf{\advCL}} }
\\
 \begin{tabular}{@{}c@{}}  
\\

 \begin{tabular}{@{\hskip 0.02in}c@{\hskip 0.02in} }
 \parbox[c]{10em}{\includegraphics[width=10em]{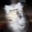}
 }  
\end{tabular} 
 \\
 \begin{tabular}{@{\hskip 0.02in}c@{\hskip 0.02in} }
 \parbox[c]{10em}{\includegraphics[width=10em]{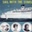}
 }  
\end{tabular}
 \\
 \begin{tabular}{@{\hskip 0.02in}c@{\hskip 0.02in} }
 \parbox[c]{10em}{\includegraphics[width=10em]{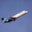}
 }  
\end{tabular}
 \\
 \begin{tabular}{@{\hskip 0.02in}c@{\hskip 0.02in} }
 \parbox[c]{10em}{\includegraphics[width=10em]{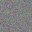}
 }  
\end{tabular}

\end{tabular} 
&
 \begin{tabular}{@{}c@{}}  
\\
 \begin{tabular}{@{\hskip 0.02in}c@{\hskip 0.02in} }
 \parbox[c]{10em}{\includegraphics[width=10em]{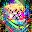}
 }  
\end{tabular} 
 \\
 \begin{tabular}{@{\hskip 0.02in}c@{\hskip 0.02in} }
 \parbox[c]{10em}{\includegraphics[width=10em]{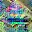}
 }  
\end{tabular}
 \\
 \begin{tabular}{@{\hskip 0.02in}c@{\hskip 0.02in} }
 \parbox[c]{10em}{\includegraphics[width=10em]{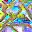}
 }  
\end{tabular}
 \\
  \begin{tabular}{@{\hskip 0.02in}c@{\hskip 0.02in} }
 \parbox[c]{10em}{\includegraphics[width=10em]{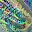}
 }  
\end{tabular}
 \\
 
\end{tabular}

&
 \begin{tabular}{@{}c@{}}  
\\

 \begin{tabular}{@{\hskip 0.02in}c@{\hskip 0.02in} }
 \parbox[c]{10em}{\includegraphics[width=10em]{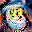}

 }  
\end{tabular} 
 \\
 \begin{tabular}{@{\hskip 0.02in}c@{\hskip 0.02in} }
 \parbox[c]{10em}{\includegraphics[width=10em]{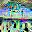}
 }  
\end{tabular}
 \\

 \begin{tabular}{@{\hskip 0.02in}c@{\hskip 0.02in} }
 \parbox[c]{10em}{\includegraphics[width=10em]{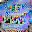}

 }  
\end{tabular}
 \\
 \begin{tabular}{@{\hskip 0.02in}c@{\hskip 0.02in} }
 \parbox[c]{10em}{\includegraphics[width=10em]{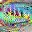}

 }  
\end{tabular}
 \\

\end{tabular} 
&
 \begin{tabular}{@{}c@{}}  
\\

 \begin{tabular}{@{\hskip 0.02in}c@{\hskip 0.02in} }
 \parbox[c]{10em}{\includegraphics[width=10em]{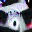}

 }  
\end{tabular} 
 \\

 \begin{tabular}{@{\hskip 0.02in}c@{\hskip 0.02in} }
 \parbox[c]{10em}{\includegraphics[width=10em]{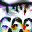}

 }  
\end{tabular}

 \\
 \begin{tabular}{@{\hskip 0.02in}c@{\hskip 0.02in} }
 \parbox[c]{10em}{\includegraphics[width=10em]{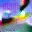}
 }  
\end{tabular}
 \\
 \begin{tabular}{@{\hskip 0.02in}c@{\hskip 0.02in} }
 \parbox[c]{10em}{\includegraphics[width=10em]{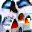}
 }  
\end{tabular}
 \\

\end{tabular} 
&
 \begin{tabular}{@{}c@{}}  
\\

 \begin{tabular}{@{\hskip 0.02in}c@{\hskip 0.02in} }
 \parbox[c]{10em}{\includegraphics[width=10em]{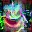}

 }  
\end{tabular} 
 \\
 \begin{tabular}{@{\hskip 0.02in}c@{\hskip 0.02in} }
 \parbox[c]{10em}{\includegraphics[width=10em]{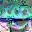}
 }  
\end{tabular}
 \\

 \begin{tabular}{@{\hskip 0.02in}c@{\hskip 0.02in} }
 \parbox[c]{10em}{\includegraphics[width=10em]{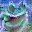}
 }  
\end{tabular}
\\
  \begin{tabular}{@{\hskip 0.02in}c@{\hskip 0.02in} }
 \parbox[c]{10em}{\includegraphics[width=10em]{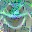}
 }  
\end{tabular}
 \\

\end{tabular} 

\end{tabular}
  \end{adjustbox}
  \vspace{-5pt}
\caption{\small{FIM visualization of neuron $502$
under CIFAR-10 using different robust training methods. Column $1$ contains different seed images to generate FIM. 
Columns 2-5 are  FIMs using models trained with different approaches.
}}
\label{fig:visualization}
\vspace*{-3mm}
\end{wrapfigure}

\vspace{-5pt}
\paragraph{Visual interpretability   of learned representations}
Furthermore, 
we demonstrate the advantage of our proposals from the perspective of model explanation,  characterized by feature inversion map (FIM) \cite{mahendran2016visualizing} of internal neurons' response.
The work \cite{boopathy2020proper,engstrom2019adversarial,kaur2019perceptually}   showed that 
model robustness offered by supervised AT and its variants enforces hidden neurons to learn    
perceptually-aligned data features through the lens of FIM.
However, it remains unclear whether or not
 \textit{self-supervised} robust pretraining is able to render explainable internal response. Following \cite{mahendran2016visualizing,engstrom2019adversarial}, we acquire FIM of the $i$th component of representation vector by solving the optimization problem 
 $
 x_{\mathrm{FIM}} = \min_{\Delta} [ f_{\theta}(x_0 + \Delta) ]_i
 $, where $x_0$ is a randomly selected seed image, and $[\cdot]_i$ denotes the $i$th coordinate of a vector. 
 Figure\,\ref{fig:visualization} shows that compared to other approaches, more similar texture-aligned  features can be acquired from a neuron's feature representation of the network trained with our method regardless of the choice of seed images.
 
\begin{figure}[htb]
\begin{center}
\vspace*{-2mm}
\begin{tabular}{ccc}
\includegraphics[width=0.3\textwidth]{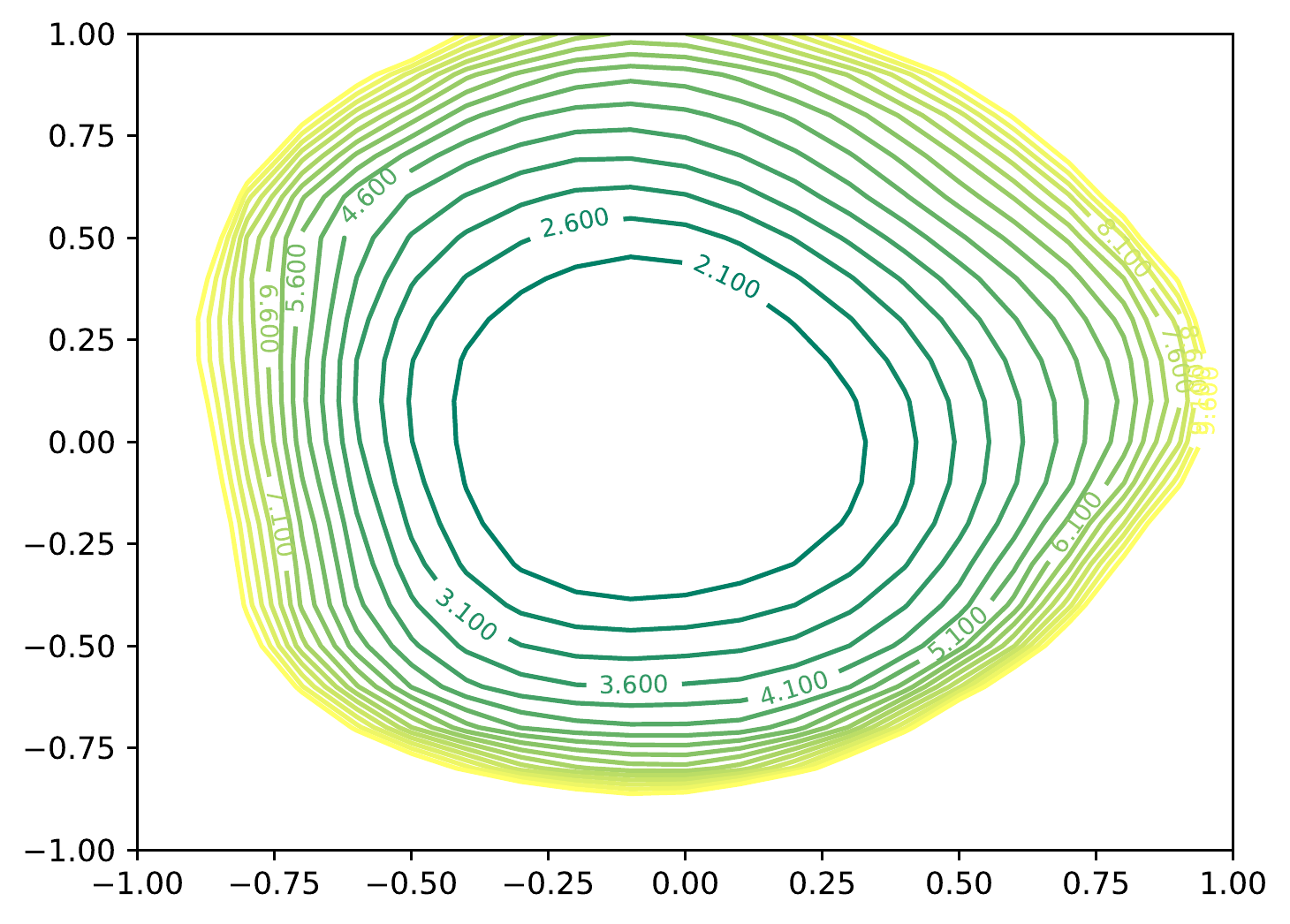} & \hspace*{-0.12in}
\includegraphics[width=0.3\textwidth]{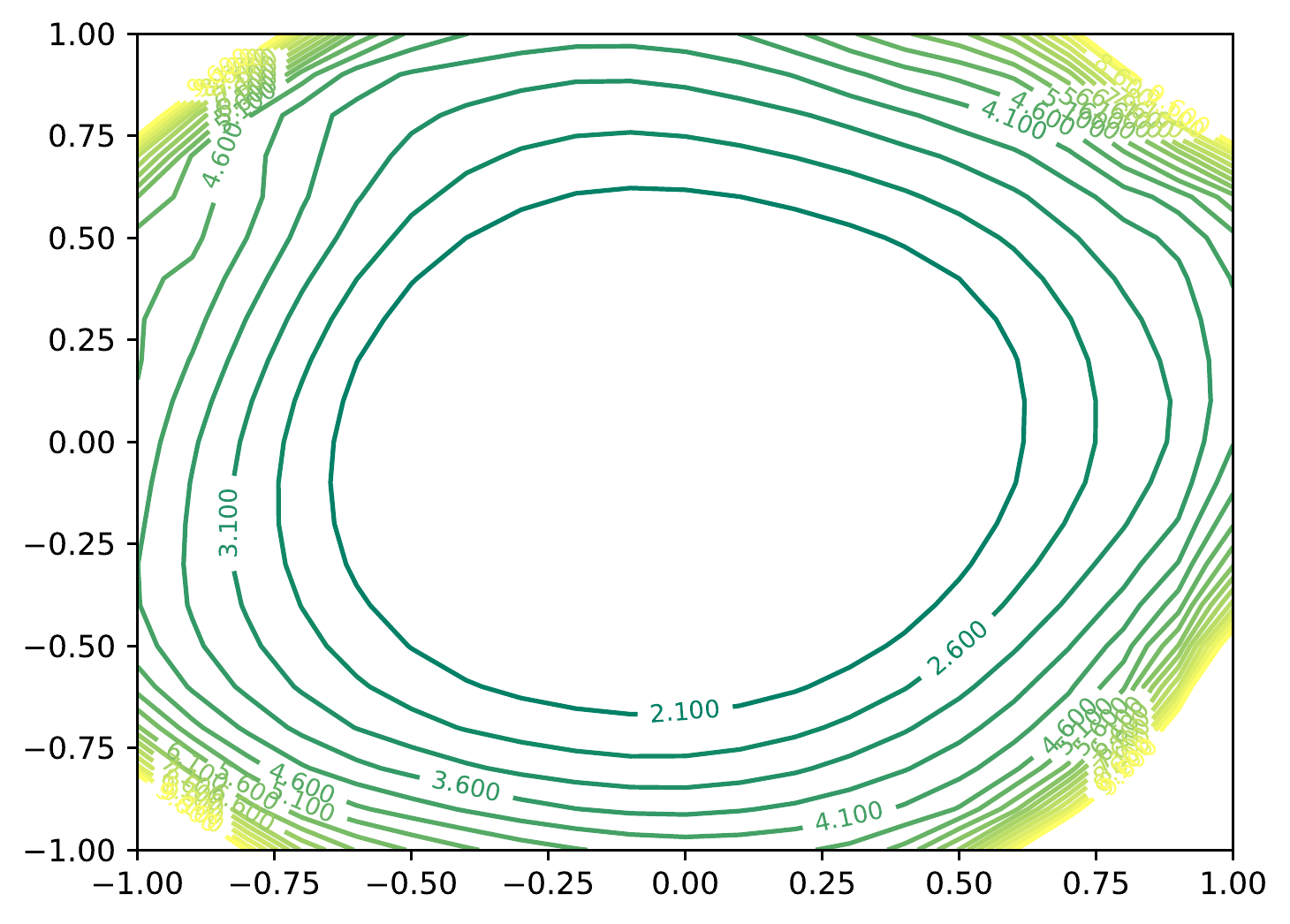} & \hspace*{-0.12in}
\includegraphics[width=0.3\textwidth]{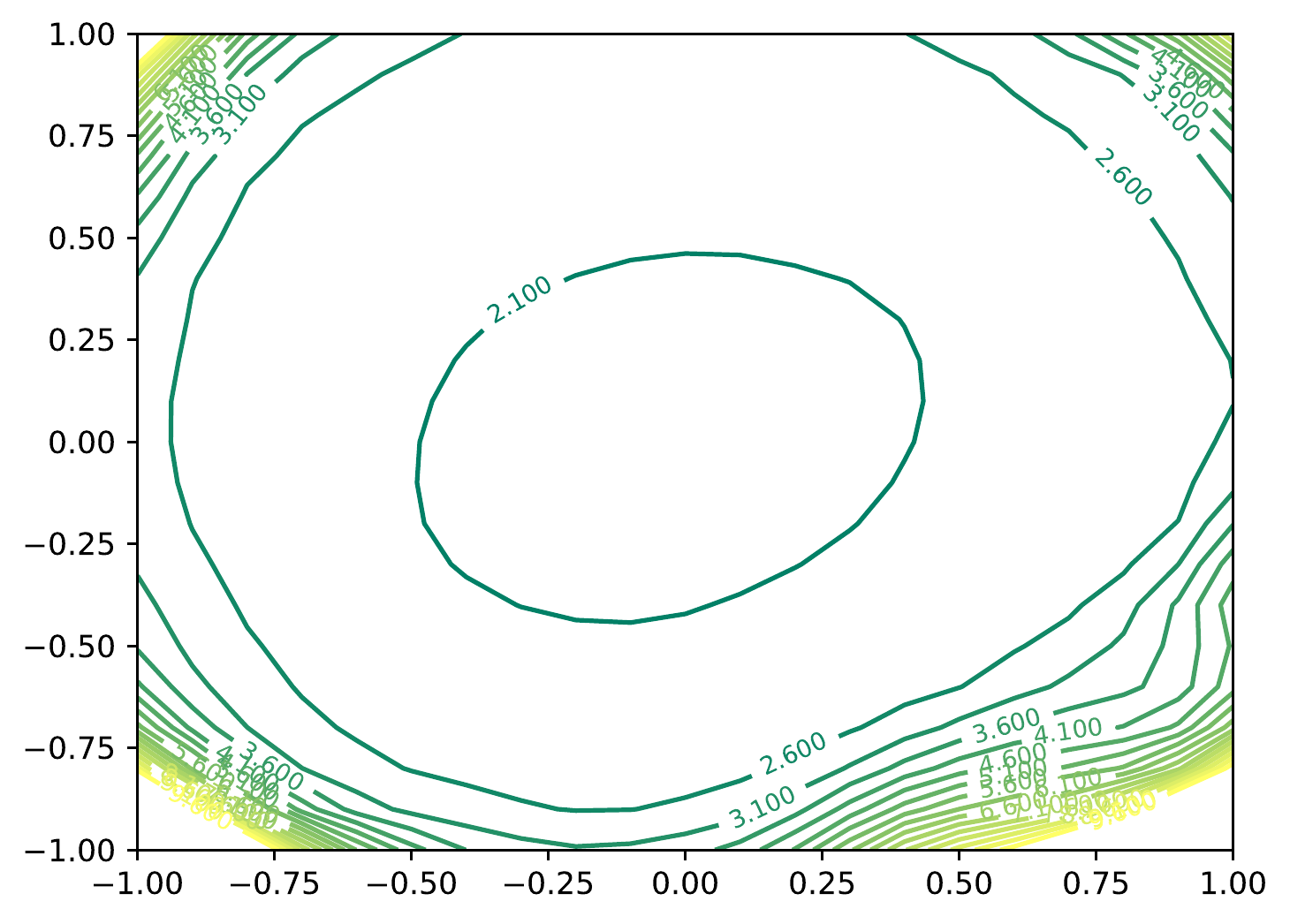}
\\
{\footnotesize (a) RoCL \cite{kim2020adversarial}} & \hspace*{-0.12in} {\footnotesize (b) ACL \cite{jiang2020robust}} & \hspace*{-0.12in} {\footnotesize (c) \advCL \bf(ours)} 
\end{tabular}
\end{center}
\vspace*{-3mm}
\caption{\footnotesize{Visualization of adversarial loss landscape w.r.t. model weights using different self-supervised pretraining methods.   {\advCL} gives a much flatter landscape than the other baselines.
}}
\label{fig:loss_landscape}
\vspace{-10pt}
\end{figure}

\vspace{-5pt}
\paragraph{Flatter loss landscape implies better transferability}
It has been shown in \cite{liu2019towards} that the flatness of loss landscape is a good indicator for superb   transferability in the pretraining + finetuning paradigm.  Motivated by that, 
Figure\,\ref{fig:loss_landscape} presents
 the adversarial loss landscape of   {\advCL} and other self-supervised pretraining approaches under SLF, 
 where the loss landscape is drawn using the method in \cite{li2017visualizing}.
 Note that instead of standard CE loss, we visualize the adversarial loss w.r.t. model weights. As we can see,  the loss for {\advCL} has a much flatter landscape around the local optima, whereas the losses for the other approaches change more rapidly. This justifies that   our proposal   has a better robustness transferability than baseline approaches. 
 
 \begin{table}[t]
 \begin{tabular}{cc}
 \begin{minipage}[t]{0.45\linewidth}
\centering
\caption{\footnotesize{Performance (RA and SA) of {\advCL} (in dark gray color) and baseline approaches on CIFAR-10, under different linear finetuning strategies: {\slf} and adversarial linear finetuning (\alf).}}
\vspace{-2mm}
\label{table:alf}
\begin{threeparttable}
\resizebox{\textwidth}{!}{
\begin{tabular}{l|c|c|c|c}
\toprule[1.2pt]\hline
\multirow{2}{*}{Method}  & \multicolumn{2}{c|}{SLF} & \multicolumn{2}{c}{ALF}  \\ 
\cmidrule{2-5}
                                                              & RA(\%)    & SA(\%)\textbf{}    & RA(\%)    & SA(\%)             \\ 
\hline
Supervised              & 44.40            & 79.77 & 46.75 & 79.06          \\

\CCGL{}RoCL\cite{kim2020adversarial}                       & \CCGL{}39.54   &\CCGL{}79.90        & \CCGL{}43.11 & \CCGL{}77.33        \\
\CCGL{}ACL\cite{jiang2020robust}                       & \CCGL{}42.87  & \CCGL{}77.88 &\CCGL{}45.40 &\CCGL{}77.71            \\
\CCG{}\bf \advCL(ours)                       & \CCG{}\bf50.45      & \CCG{}\bf80.85  & \CCG{}\bf52.01       & \CCG{}\bf79.39           \\
\hline\bottomrule[1.2pt]
\end{tabular}}
\end{threeparttable}
\vspace{1pt}
\caption{\footnotesize{Performance (RA and SA) of {\advCL} using different contrastive views  setups. ResNet-18 is  the backbone network, CIFAR-10 is the dataset, and SLF is used for classification.
}
}
 \vspace{-4pt}
\label{table:view}
\begin{threeparttable}
\resizebox{.99\textwidth}{!}{
\begin{tabular}{l|c|c}
\toprule[1.2pt]\toprule
Contrastive Views & RA(\%) & SA(\%)  \\
\hline
 $\tau_1(x)+\delta_1, \tau_2(x)$                &     42.12          &      77.07                      \\
$\tau_1(x)+\delta_1, \tau_2(x)+\delta_2$        &       42.48        &      73.12                  \\
$\tau_1(x)+\delta_1, \tau_2(x)+\delta_2, \tau_1(x), \tau_2(x)$ &        43.51                    &     74.22          \\
 ${x+\delta, \tau_1(x), \tau_2(x)}$               &    \bf 50.19         &       \bf 80.17                  \\
\midrule

\rowcolor{GrayL}$x+\delta, \tau_1(x), \tau_2(x), x_{\mathrm{l}}$          &        49.51       &            79.83                \\
\rowcolor{GrayL}$x+\delta, \tau_1(x), \tau_2(x), x_{\mathrm l}, x_{\mathrm{h}}$      &       50.03        &          80.14                \\
\rowcolor{GrayL} ${x+\delta, \tau_1(x), \tau_2(x), x_{\mathrm{h}}}$         &        \bf 50.45       &           \bf 80.85               \\
\bottomrule\bottomrule[1.2pt]
\end{tabular}}
\end{threeparttable}
\end{minipage}
 &
\begin{minipage}[t]{0.49\linewidth}
\centering
\caption{\small{Performance (RA and SA) of {\advCL} using various pretrained models $f_{\mathrm{pre}}$ and cluster numbers $K$  in {\CFit}, as well as the baseline w/o using {\CFit}. The setup of $f_{\mathrm{pre}}$ is specified by the training method (supervised training or SimCLR) and training dataset (ImageNet or CIFAR-10). {\advCL} is implemented using unlabeled data from CIFAR-10 under ResNet-18,
together with {\slf} over the acquired feature encoder for supervised CIFAR-10 classification.}
}
 \vspace*{0.5mm}
\label{tab:selflabeling}
\begin{threeparttable}
\resizebox{\textwidth}{!}{
\begin{tabular}{c|c|c|c}
\toprule[1.2pt]\toprule
    \begin{tabular}[c]{@{}c@{}}{$f_{\mathrm{pre}}$ setup:}\\(dataset, training) \end{tabular}
 & Cluster number $K$       &  RA (\%) & SA (\%)  \\   
\midrule
N/A & \begin{tabular}[c]{@{}c@{}}
w/o {\CFit} \end{tabular} & 48.89 & 77.73
\\
\midrule
\multirow{2}{*}{\begin{tabular}[c]{@{}c@{}}(CIFAR-10, SimCLR)\end{tabular}} & 10  & 50.10                                                                 & 80.34                             \\
 & 100  & 49.21     & 79.52     \\ 
     \midrule
\multirow{2}{*}{\begin{tabular}[c]{@{}c@{}}(ImageNet, supervised)\end{tabular}} & 10  & 50.16                                                                 & 78.27                             \\
 & 100  & 49.27     & 78.08     \\ 
\midrule
\multirow{6}{*}{\begin{tabular}[c]{@{}c@{}}(ImageNet, SimCLR)\end{tabular}} & 2  & 50.09                                                                 & 79.72                             \\
 & 10  & 50.12     & 79.93     \\ 
  & 50  & 49.27     & 79.55     \\ 
   & 100  & 49.16     & 79.07     \\ 
    & 500  & 49.03     & 78.96     \\
    \cmidrule{2-4}
    & \bf Ensemble  & \bf 50.45     & \bf 80.85     \\ 
\bottomrule\bottomrule[1.2pt]
\end{tabular}}
\end{threeparttable}
\end{minipage}
\end{tabular}
\end{table}

 \subsection{Ablation studies}
 \vspace{-5pt}
 \paragraph{Linear finetuning types}
We first study the robustness difference when different linear finetuning strategies: \textit{Standard} linear finetuning (SLF) and \textit{Adversarial} linear finetuning (ALF) are applied. 
Table\,\ref{table:alf} shows the performance of models trained with different pretraining methods. 
As we can see, our {\advCL} achieves the best performance under both linear finetuning settings and outperforms baseline approaches in a large margin.  We also note  the performance gap between SLF and ALF induced by our proposal {\advCL} is much smaller than other approaches, and {\advCL} with {\slf} achieves much better performance than baseline approaches with {\alf}. This indicates that the representations learned by {\advCL} is already sufficient to yield satisfactory robustness.

\vspace{-5pt}
 \paragraph{View selection setup}
We illustrate how different choices of contrastive views influence the robustness performance of {\advCL} in Table \ref{table:view}. The first 4 rows study the effect of different types of adversarial examples in contrastive views, and our proposed 3-view contrastive loss \eqref{eq: atk_3view} significantly outperforms the other baselines, as shown in row 4. The rows in gray show the performance of further exploring different image frequency components \eqref{eq: LHFC} as different contrastive views. It is clear that the use of HFC leads to the best overall performance, as shown in the last row.

\vspace{-5pt}
\paragraph{Supervision stimulus setup}
We further study the performance of {\advCL} using different supervision stimulus. Specifically, we vary the pretrained model for $f_{\mathrm{pre}}$ and pseudo cluster number $K$ when training {\advCL} and summarize the results in Table \ref{tab:selflabeling}. The results demonstrate that adding the supervision stimulus could boost the performance of {\advCL}.
{We also observe that the best result comes from $f_{\mathrm{pre}}$ pretrained on Imagenet using SimCLR. This is because such representations could generalize better. Moreover, the ensemble scheme over pseudo label categories $K \in \{2,10,50,100,500\}$ yields better results than using a single number of clusters. The ensemble scheme also makes {\advCL} less sensitive to the actual number of labels for the training dataset.}

\vspace{-6pt}
\section{Conclusion}
\vspace{-1pt}
In this paper, we study the good practices in making contrastive learning robust to adversarial examples. We show that adding perturbations to original images and high-frequency components are two beneficial factors. We further show that proper supervision stimulus could improve model robustness. Our proposed approaches  can achieve state-of-the-art   robust accuracy as well as standard accuracy using just   standard linear finetuning. Extensive experiments involving quantitative and qualitative analysis have also been made  not only to demonstrate the effectiveness of our proposals but also to rationalize why  it yields superior performance.
Future works could be done to improve the scalability of our proposed self-supervised pretraining approach to very large datasets and models to further boost robust transferabilty across datasets.

\clearpage
\bibliographystyle{IEEEbib}
\bibliography{refs, ref_SL_adv, ref_SL_fair_self}

\clearpage
\appendix
\appendixpage


\setcounter{figure}{0}
\makeatletter 
\renewcommand{\thefigure}{A\@arabic\c@figure}
\makeatother
\setcounter{table}{0}
\renewcommand{\thetable}{A\arabic{table}}
\setcounter{algorithm}{0}
\renewcommand{\thealgorithm}{A\arabic{algorithm}}

This supplementary material provides additional implementation details and experimental results.

\vspace{-2mm}
\section*{A. Discussion and Broader Impact}
\vspace{-2mm}

In this paper we propose a powerful framework, {\advCL}, which could preserve robustness from pretraining to finetuning, and we empircally show that the light-weight standard linear finetuning is already sufficient to give us comparable performance to the computational-expensive adversarial full finetuning.
We don't think  our work would have negative societal impacts.
The potential broader impact of our work is that, with the help of our proposed pretraining design paradigm, neural models could preserve adversarial robustness using lightweight linear finetuners, which could be deployed to embodied systems and can make real-time applications more secure and trustworthy on mobile devices.

\vspace{-2mm}
\section*{B. Implementation Details}
\vspace{-2mm}
\paragraph{Pretraining Details}
We list implementation details for {\advCL} pretraining   in this section. We use SGD optimizer with batch size=$512$, initial learning rate=$0.5$, momentum=$0.9$, and weight decay=$0.0001$ to train the network for $1000$ epochs. We use cosine learning rate decay during training, and we use the first $10$ epochs to warm-up learning rate from $0.01$ to $0.5$. The temperature parameter in contrastive loss $\ell_{\mathrm{CL}}$ is set to $t=0.5$, and the regularization parameter for the cross-entropy term with pseudo labels in Eq.(10) is set to $\lambda=0.2$.
All experiments are performed on 4 NVIDIA TITAN Xp GPUs.

The augmentation set $\mathcal{T}$ for pretraining consists of random cropping with scale $0.2$ to $1$, random horizontal flip, random color jittering and random grayscale. We provide the pseudo code for implementing $\mathcal{T}$ here in PyTorch in Algorithm \ref{alg:code}.

\paragraph{Finetuning Details} 
For \slf, the encoder parameters $f_\theta$ are fixed and the linear classifier is trained for 25 epochs using SGD with batch size=$512$, initial learning rate=$0.1$, momentum=$0.9$, and weight decay=$0.0002$. The learning rate is decreased to $0.1\times$ at epoch $15, 20$. 
For \alf, we use $10$-step $\ell_\infty$ PGD attack with $\epsilon=8/255$ to generate adversarial perturbations during training. The encoder parameters $f_\theta$ are fixed and the linear classifier is trained for 25 epochs using SGD with batch size=$512$, initial learning rate=$0.1$. The learning rate is decreased to $0.1\times$ at epoch $15, 20$. 
For \aff, following the settings in \cite{jiang2020robust}, we also use $10$-step $\ell_\infty$ PGD attack with $\epsilon=8/255$ to generate adversarial perturbations during training, and train the entire network parameters $f_\theta$ and the linear classifier with trades loss for $25$ epochs with initial learning rate of $0.1$ which decreases to $0.1\times$ at epoch $15, 20$. 
We report the AA, RA and SA for the best possible model for every method under every setting.

\paragraph{{\TBN}: Customized batch normalization}
It has recently been shown in \cite{jiang2020robust,galloway2019batch,pang2020bag} that batch normalization (BN) could play a vital role in robust training with `mixed' normal and adversarial data.
Thus, a careful study on the BN strategy of {\advCL} is needed, since \textit{two types of adversarial perturbations}  are generated in 
Eq.(10) w.r.t. different adversary's goals, maximizing the CL loss ($\delta$) vs. maximizing the CE loss ($\delta_{\mathrm{ce}}$). 
Thus, to fit different adversarial data distributions, we introduce two BNs, each of which 
corresponds to one adversary type. 
Besides, we use the other BN for \textit{normally transformed data}, i.e., $(\tau_1(x), \tau_2(x), x_{\mathrm{h}})$.
Compared with existing work \cite{jiang2020robust,galloway2019batch,pang2020bag} that used 
$2$ BNs (one for adversarial data and the other for benign data), our proposed {\advCL} calls for triple BNs ({\TBN}).


\section*{C. Performance Summary under AutoAttack}
\begin{figure}[t]
 \begin{tabular}{cc}
 \begin{minipage}{0.44\textwidth}
\begin{algorithm}[H]
\caption{Pseudocode of Augmentation $\mathcal{T}$ in PyTorch.}
\label{alg:code}
\definecolor{codeblue}{rgb}{0.25,0.5,0.5}
\lstset{
  backgroundcolor=\color{white},
  basicstyle=\fontsize{8.2pt}{8.2pt}\ttfamily\selectfont,
  columns=fullflexible,
  breaklines=true,
  captionpos=b,
  commentstyle=\fontsize{8.2pt}{8.2pt}\color{codeblue},
  keywordstyle=\fontsize{8.2pt}{8.2pt},
}
\begin{lstlisting}[language=python]
transform = transforms.Compose([
    # random cropping
    transforms.RandomResizedCrop(size=32, scale=(0.2, 1.)),
    # random horizontal flip
    transforms.RandomHorizontalFlip(),
    # random color jittering
    transforms.RandomApply([
        transforms.ColorJitter(0.4, 0.4, 0.4, 0.1)
    ], p=0.8),
    # random grayscale
    transforms.RandomGrayscale(p=0.2),
    transforms.ToTensor(),
])
\end{lstlisting}
\end{algorithm}
\end{minipage} &
\begin{minipage}{0.52\textwidth}
\includegraphics[width=20em, height=14em]{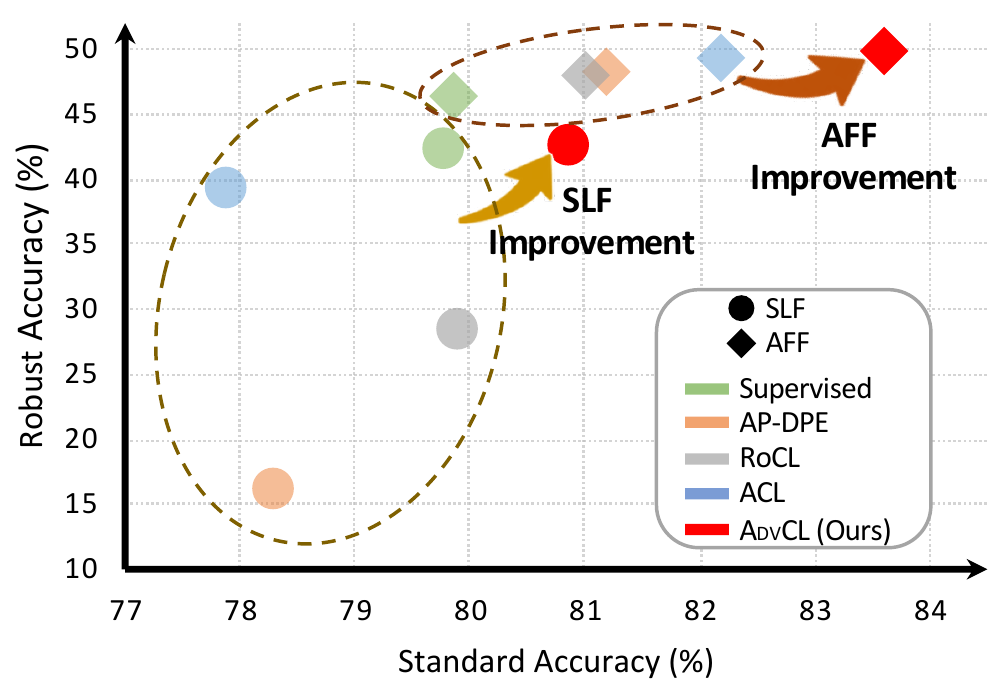}
\vspace{-10pt}
\caption{\footnotesize{Performance of various robust pretrainig methods on CIFAR-10.
Upper-right indicates better performance w.r.t. standard accuracy and robust accuracy (under Auto-Attack with $8/255$ $\ell_\infty$-norm perturbation strength). 
}}
\label{fig:teaser_aa}
\end{minipage}
\end{tabular}
\end{figure}

In analogy to Figure\,1 of the main paper, Figure\,\ref{fig:teaser_aa} 
 shows the performance comparison of {\advCL} and baseline approaches with respect to (w.r.t) Auto-Attack Accuracy (AA) and Standard Accuracy (SA) both under {\slf} and {\aff} on CIFAR-10.
We compare our proposed approaches with self-supervised pretraining baselines: AP-DPE \cite{chen2020adversarial}, RoCL \cite{kim2020adversarial}, ACL \cite{jiang2020robust} and supervised adversarial training (AT) \cite{madry2017towards}.
Upper-right indicates better performance w.r.t. standard accuracy and robust accuracy (under Auto-Attack with $8/255$ $\ell_\infty$-norm perturbation strength). Colors represents pretraining approaches, and shapes represent finetuning settings. 
{Circles} (\ding{108}) indicates \textit{Standard Linear Finetuning} (SLF), and {Diamonds} (\ding{117}) indicates \textit{{Adversarial Full Finetuning}} (AFF). As we can see, our proposed approach (\advCL, red circle/diamond) has the best performance across both {\slf} and {\aff} settings and outperforms all baseline approaches significantly.

\section*{D. Comparing SLF with ALF}
\begin{figure}[htb]
\begin{center}
\vspace*{-2mm}
\includegraphics[width=.98\textwidth,height=!]{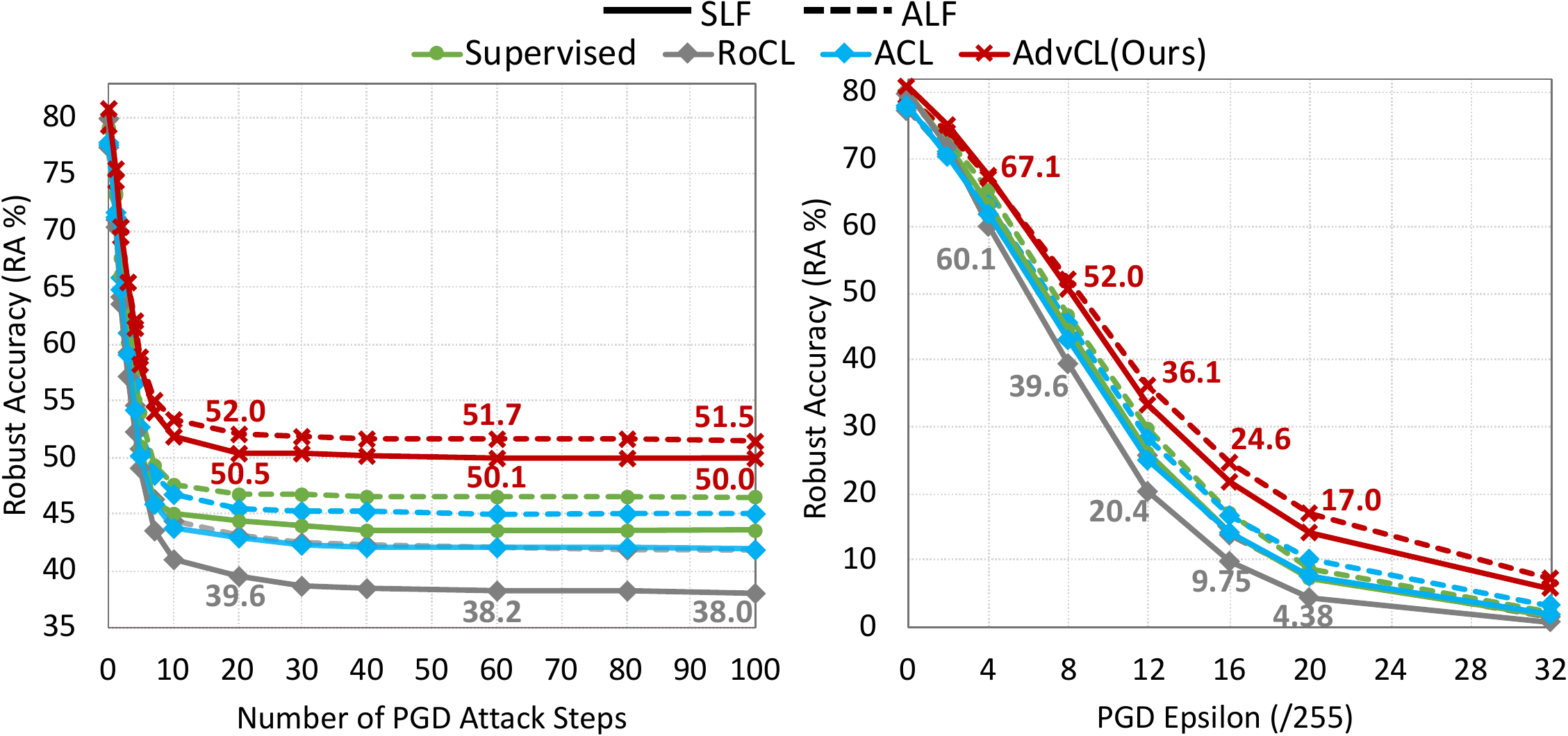}
\end{center}
\vspace*{-3mm}
\caption{\footnotesize{Robust accuracy (RA) of different pretraining approaches under various PGD attacks on CIFAR-10. SLF or ALF is applied to the pretrained model. Different colors represent different pretraining approaches, and different line types represent different linear finetuning approaches (solid line for {SLF} and dash line for ALF). Our proposed approach ({\advCL}, red  line) outperforms the baseline approaches in a non-trivial margin under all attack settings.
}}
\label{fig:attack_adv}
\vspace{-3pt}
\vspace{-5pt}
\end{figure}

Here we futher compare the performance gap of Standard Linear Finetuning(SLF) and Adversarial Linear Finetuning (ALF) with different pretraining approaches, by attacking the final model with various PGD attacks on CIFAR-10.
The summarized results are in Figure \ref{fig:attack_adv}.  Different colors represent different pretraining approaches, and different line types represent different linear finetuning approaches. (\textbf{Solid line} for {SLF} and \textbf{dash line} for ALF). As we can see,
{\advCL}+{\slf} is already sufficient to outperform all baseline approaches under {\alf}. When the latter is applied to {\advCL}, robustness is further improved by a small margin. This is different from baseline methods, where
 using ALF can boost the performance  by a large margin. 
This phenomena justifies that the representations learned by {\advCL} is more robust so that a standard finetuned linear classifier can already make the whole model robust, while the baseline approaches will need the linear classifier to be trained adversarially to obtain more satisfactory results.  
To the best of our knowledge, our approach is the only self-supervised pretraining approach that can outperform the supervised AT baseline under both {\slf} and {\alf}.

\section*{E. Experiments on More Vision Datasets}
We conducted more experiments on two in-domain settings: SVHN and TinyImageNet, and two cross-domain settings: SVHN$\xrightarrow{}$STL10 and TinyImageNet$\xrightarrow{}$STL10. We compare the performance of \name with that of the other baseline methods in the Standard Linear Finetuning setup. The results are summarized in Table~\ref{table:exp-add-1} and \ref{table:exp-add-2}.

As we can see, our proposed \name outperforms the other self-supervised and supervised adversarial training (AT) baselines in most cases, except for the in-domain TinyImageNet case on SA. However, the ~$0.2\%$ drop of SA corresponds to a more significant RA improvement of ~$3.4\%$. These results further justify the effectiveness of our proposal.

\section*{F. Running Time Comparison}
\paragraph{Different Finetuning Method}
Adversarial full finetuning (AFF) is much more computationally intensive than standard linear finetuning (SLF) per epoch. In Table~\ref{table:time-finetune}, we list the detailed training time comparison between SLF and AFF. As we can see, AFF takes $24\times$ more training time than SLF. That is because AFF has to call for the min-max optimization (multiple inner-level maximization iterations needed per outer-level minimization step) to preserve model robustness. 

\paragraph{Different Pretraining Method}
We also demonstrate the training time costs of different pretraining methods in Table~\ref{tabl:time-pretrain}. We can make several observations:
\begin{enumerate}

\item
Self-supervision-based pretraining methods take more time than the supervised AT method since self-supervised pretraining approaches typically require more epochs than fully supervised methods to converge.
\item
In the self-supervision-based pretraining approaches, AP-DPE\cite{chen2020adversarial} takes the highest computation cost as it resorts to a complex min-max-based ensemble training recipe. Compared to the contrastive learning-based baselines (RoCL and ACL), ours (\advCL) takes higher computation cost. This is because: (a) the contrastive loss of \name takes more than two image views, and (b) \name calls an additional pseudo supervision regularization. However, the pretraining procedure can often be conducted offline. Thus, the finetuning efficiency (via SLF) still makes \name advantageous over the other baselines to preserve model robustness from pretraining to downstream tasks.
\end{enumerate}

 \begin{table}[t]
 \begin{tabular}{cc}
 \begin{minipage}[t]{0.45\linewidth}
\centering
\caption{\footnotesize{Training time comparison for different finetuning approaches over a pretrained model.}}
\vspace{-2mm}
\label{table:time-finetune}
\begin{threeparttable}
\resizebox{0.99\textwidth}{!}{
\begin{tabular}{c|c}
\toprule[1.2pt]\hline
\begin{tabular}[c]{@{}c@{}}{Finetuning Method} \end{tabular}
 & \begin{tabular}[l]{@{}c@{}}Running Time \\(per epoch $\times$ epochs)\end{tabular}   \\
\midrule
Standard Linear (\slf) & 5.58s$\times$25\\
Adversarial Full (\aff) & 136.08s $\times$25\\
\hline\bottomrule[1.2pt]
\end{tabular}}
\end{threeparttable}
\vspace{1pt}
\caption{\small{Training time comparison for different pretraining approaches.}
}
 \vspace*{-1mm}
\label{tabl:time-pretrain}
\begin{threeparttable}
\resizebox{0.99\textwidth}{!}{
\begin{tabular}{c|c}
\toprule[1.2pt]\toprule
   \begin{tabular}[c]{@{}c@{}}{Finetuning Method} \end{tabular}
 & \begin{tabular}[l]{@{}c@{}}Running Time \\(per epoch $\times$ epochs)\end{tabular}   \\
\midrule
Supervised AT & 69.55s$\times$200\\
AP-DPE\cite{chen2020adversarial} & 6979.58s$\times$150\\
RoCL\cite{kim2020adversarial} & 123.15s$\times$1000\\
ACL\cite{jiang2020robust} & 126.8s$\times$1000\\
\advCL(ours) & 377.89s$\times$1000\\

\bottomrule\bottomrule[1.2pt]
\end{tabular}}
\end{threeparttable}

\end{minipage}
 &
\begin{minipage}[t]{0.49\linewidth}
\centering

\caption{\footnotesize{Performance of \advCL, compared with baselines, in terms of RA and SA on SVHN and TinyImageNet, under Standard Linear Finetuning.}}.

\vspace{-4mm}
\label{table:exp-add-1}
\begin{threeparttable}
\resizebox{0.9\textwidth}{!}{
\begin{tabular}{c|cc|cc} 
\toprule[1.2pt]\toprule
\multirow{2}{*}{\begin{tabular}[c]{@{}c@{}}Pretraining\\Method\end{tabular}}        & \multicolumn{2}{c|}{SVHN} & \multicolumn{2}{c}{TinyImageNet}  \\
\cmidrule{2-5}
                                                                         &                                                                                          RA(\%)  & SA(\%) & RA(\%)  & SA(\%)             \\ 
\midrule
Supervised & 41.03 & 91.13 & 16.03 & 42.73     \\
ACL\cite{jiang2020robust} & 39.34 & 89.64 & 17.25 & 41.33 \\
\name(ours) & 45.15 & 92.85 & 19.39 & 42.50  \\
\bottomrule\bottomrule[1.2pt]
\end{tabular}}

 \vspace*{1mm}
\end{threeparttable}

\caption{\footnotesize{Cross-dataset performance of \advCL, compared with baselines, in terms of RA and SA, under Standard Linear Finetuning.}}.

\vspace{-3mm}
\label{table:exp-add-2}
\begin{threeparttable}
\resizebox{0.9\textwidth}{!}{
\begin{tabular}{c|cc|cc} 
\toprule[1.2pt]\toprule
\multirow{2}{*}{\begin{tabular}[c]{@{}c@{}}Pretraining\\Method\end{tabular}}        & \multicolumn{2}{c|}{SVHN$\xrightarrow[]{}$STL-10} & \multicolumn{2}{c}{TinyImgNet$\xrightarrow[]{}$STL-10}  \\
\cmidrule{2-5}
                                                                         &                                                                                          RA(\%)  & SA(\%) & RA(\%)  & SA(\%)             \\ 
\midrule
Supervised & 13.81 & 38.97 & 24.31 & 60.93     \\
ACL\cite{jiang2020robust} & 15.24 & 38.53 & 28.52 & 59.64 \\
\name(ours) & 20.51 & 40.73 & 32.95 & 62.08  \\
\bottomrule\bottomrule[1.2pt]
\end{tabular}}

\end{threeparttable}
\end{minipage}
\end{tabular}
\end{table}

\section*{G. Robust Transfer Learning}
\begin{figure}[t]
\begin{center}
\includegraphics[width=.98\textwidth,height=!]{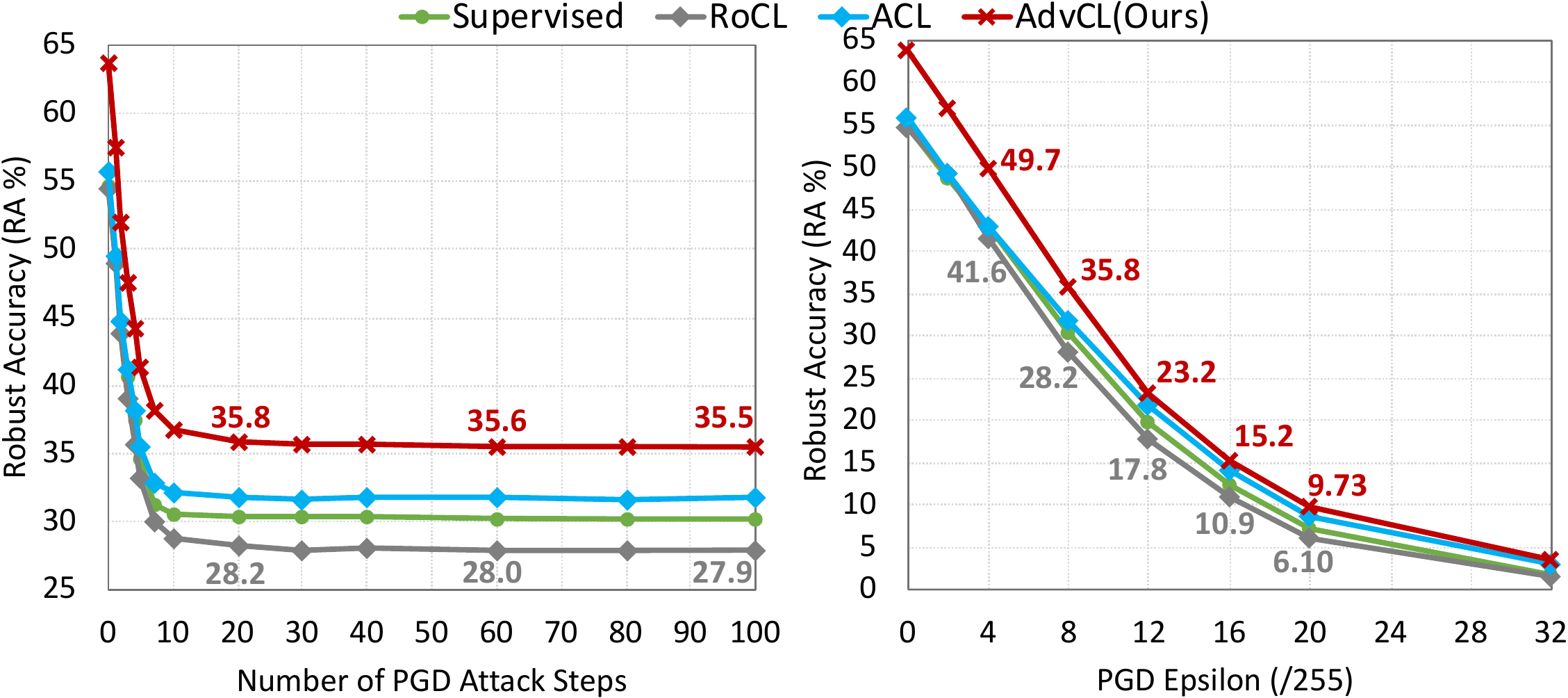}
\end{center}
\vspace*{-3mm}
\caption{\footnotesize{Robust accuracy (RA) of different pretraining approaches under various PGD attacks when transferring from CIFAR-10 to STL-10. SLF is applied to the pretrained model. Different colors represent different pretraining approaches. Our proposed approach ({\advCL}, red) outperforms the baseline approaches in a non-trivial margin.
}}
\label{fig:attack_stl10}
\vspace{-3pt}
\vspace{-5pt}
\end{figure}

We also evaluate the model performance under various PGD attacks when transferring from CIFAR-10 to STL-10. Here SLF is applied to the pretrained model. We summarize the performance in Figure \ref{fig:attack_stl10},  where different colors represent different pretraining approaches. As we can see, {\advCL} achieves the best performance under all attack settings and outperform baseline methods significantly.

\section*{H. Ablation Studies on CIFAR-100}

\begin{figure}[t]
\begin{center}
\includegraphics[width=.98\textwidth,height=!]{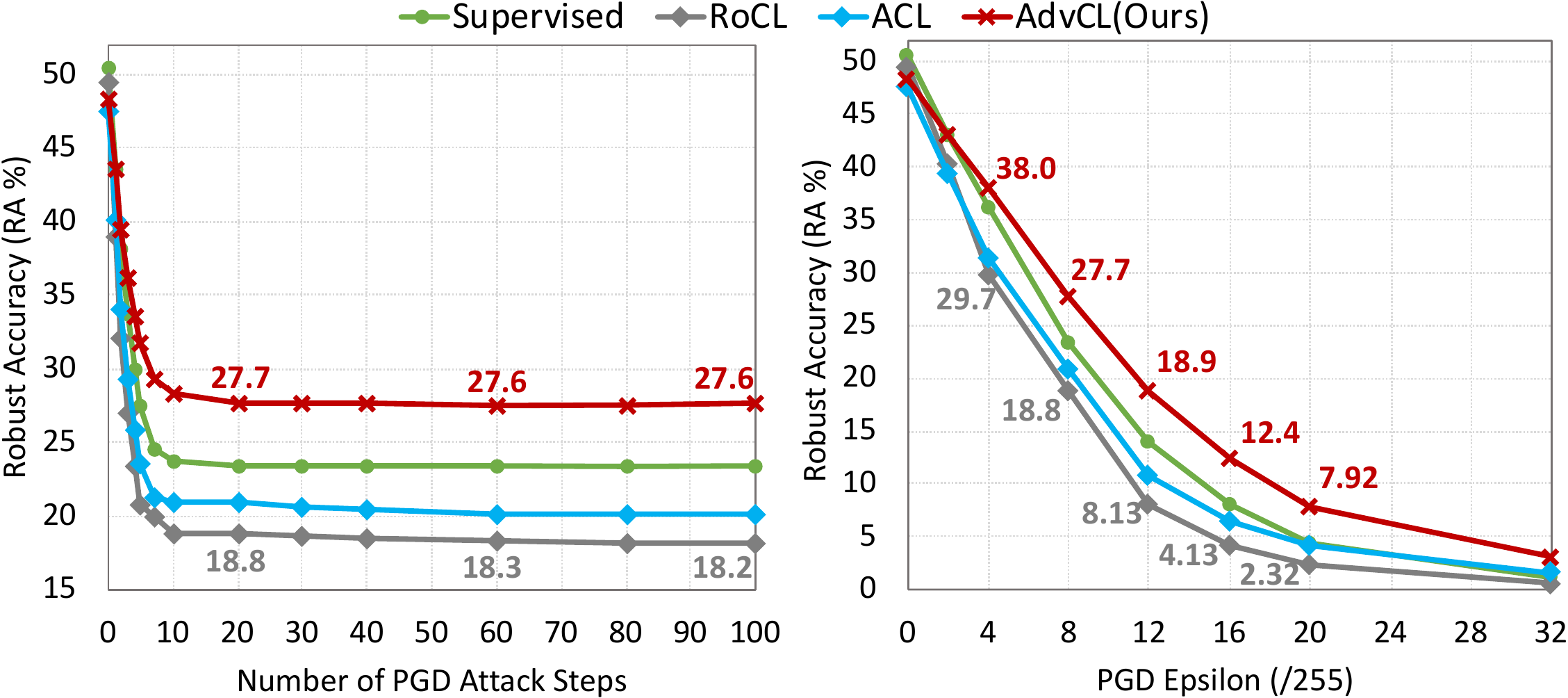}
\end{center}
\vspace*{-3mm}
\caption{\footnotesize{Robust accuracy (RA) of different pretraining approaches under various PGD attacks on CIFAR-100. SLF is applied to the pretrained model. Different colors represent different pretraining approaches. Our proposed approach ({\advCL}, red) outperforms the baseline approaches in a non-trivial margin.
}}
\label{fig:attack_cifar100}
\vspace{-3pt}
\end{figure}

\paragraph{Various attack strengths}
We also evaluate the models trained with different pretraining approaches under various PGD attacks on CIFAR-100, to further justify whether there exists the issue of obfuscated gradients on CIFAR-100. The summarized results are shown in Figure \ref{fig:attack_cifar100}. As the results suggest, our {\advCL} outperforms  baseline approaches for most of the cases, especially when the attack becomes stronger with higher step or epsilon.

\end{document}